\documentclass[authoryear,3p,times,twocolumn]{elsarticle}
\usepackage{amsthm}
\usepackage{url}

\usepackage{footnote}
\usepackage{amsmath}
\usepackage{algpseudocode}

\usepackage[ruled, vlined, linesnumbered, resetcount]{algorithm2e}
\usepackage{subfig}
\usepackage{placeins}

\usepackage[normalem]{ulem}
\usepackage{enumitem}
\usepackage{multirow}
\useunder{\uline}{\ul}{}

\usepackage{amssymb}
\usepackage{graphicx}

\usepackage{subfig}
\usepackage{multirow}

\journal{}

\begin{document}

\begin{frontmatter}



\title{Handling Adversarial Concept Drift in Streaming Data}


\author[tj]{Tegjyot Singh Sethi \corref{cor1}}
\ead{tegjyotsingh.sethi@louisville.edu}
\author[mk]{Mehmed Kantardzic}
\ead{mehmedkantardzic@louisville.edu}


\address[tj]{Data Mining Lab, University of Louisville, Louisville, USA }
\cortext[cor1]{Corresponding author.}

\begin{abstract}

Classifiers operating in a dynamic, real world environment, are vulnerable to adversarial activity, which causes the data distribution to change over time. These changes are traditionally referred to as concept drift, and several approaches have been developed in literature to deal with the problem of drift handling and detection. However, most concept drift handling techniques, approach it as a domain independent task, to make them applicable to a wide gamut of reactive systems. These techniques were developed from an adversarial agnostic perspective, where they are naive and assume that drift is a benign change, which can be fixed by updating the model. However, this is not the case when an active adversary is trying to evade the deployed classification system. In such an environment, the properties of concept drift are unique, as the drift is intended to degrade the system and at the same time designed to avoid detection by traditional concept drift detection techniques. This special category of drift is termed as adversarial drift, and this paper analyzes its characteristics and impact, in a streaming environment. A novel framework for dealing with adversarial concept drift is proposed, called the \textit{Predict-Detect} streaming framework. This framework uses adversarial forethought, and incorporates the context of classification into the drift detection task, to provide leverage in dynamic-adversarial domains. Experimental evaluation of the framework, on generated adversarial drifting data streams, demonstrates that this framework is able to provide reliable unsupervised indication of drift, and is able to recover from drifts swiftly. While traditional partially labeled concept drift detection methodologies fail to detect adversarial drifts, the proposed framework is able to detect such drifts and operates with $<$6\% labeled data, on average. Also, the framework provides benefits for active learning over imbalanced data streams, by innately providing for feature space honeypots, where minority class adversarial samples may be captured. The framework provides for an application independent, distribution independent, incremental, and semi supervised system for continuously dealing with adversarial activity at test time, and provides a generic way for implementing reactive security to classification based systems. 

\end{abstract}


\end{frontmatter}


\section{Introduction}
\label{sec:introduction}

Data in dynamic real world environments are characterized by non-stationarity. The changes in the data distribution, called concept drift, can cause the learned model to drop in predictive performance, over time \citep{gama2004learning,vzliobaite2010learning}. It is therefore essential to detect and handle drifts swiftly, to continue using the predictive capabilities of the model. Adversarial drift is a special kind of concept drift, where the changes in the data distribution are targeted towards affecting the characteristics of one class of samples (i.e., the \textit{Malicious} class). The adversary starts by learning the behavior of the defender's classifier model, using crafted probes, and then exploits this information to generate attack samples, to evade classification \citep{barreno2010security,tsethi2016,sethi2017data,biggio2013evasion}. These attacks leads to a change in the distribution of the data, at test time, and also leads to a drop in the prediction capabilities of the defender's model. From the perspective of streaming data mining, we refer to such changes in the data distribution at test time, as \textit{Adversarial Drifts} \citep{kantchelian2013approaches}. For a classifier operating in an adversarial environment, it is essential to be able to deal with such data dynamics, and to adapt the model, so as to be of any practical long term use.

\begin{figure*}[t]
  \centering
  \includegraphics[width=0.75\linewidth]{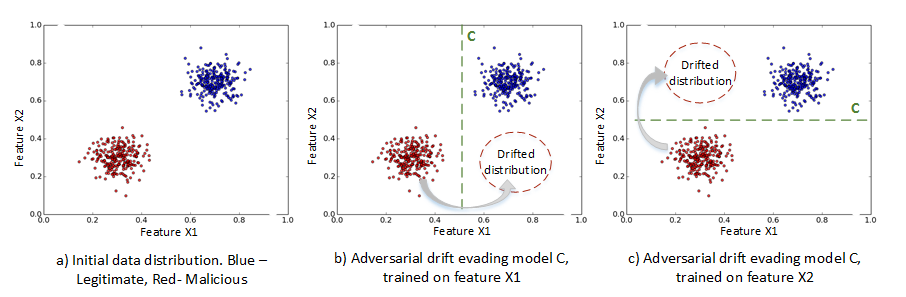}
   \caption{Illustration of adversarial drift, as a function of the defender's classifier model $C$.}
  \label{fig:adverarial_drif_function}
\end{figure*}

Adversarial drift is a special type of concept drift. The main characteristics of adversarial drift which distinguishes it from traditional concept drift are: a) The drift is a result of changes to the malicious class samples only, b) The drift is a function of the deployed classifier model, as the adversary learns and gains information about it, before trying to evade it \citep{barreno2006can}, and c) The drift is always targeted towards subverting the deployed classifier (i.e., it is relevant only if it leads to a drop in the performance of the deployed model) \citep{tsethi2016}. The dependent nature of adversarial drift is shown in Figure~\ref{fig:adverarial_drif_function}, where the deployed classifier $C$ dictates the possibility of adversarial drifts in the data space. The figures b) and c) demonstrate adversarial drifts, which are caused by an attacker trying to subvert $C$. The two scenarios are a result of the different defender models, which the adversary is trying to learn and circumvent. The nature of the drift is dependent on the choice for $C$, and as such the model designer has a certain degree of control over the possible space of drifts, at test time. 

The detection of drifts is often carried out by supervised approaches, which continuously monitor the predictive performance of the stream of data, flowing into the system  \citep{goncalves2014comparative}. However, this is not a practical solution in a streaming environment, as human expertise in the form of labeled data, is often expensive and time taking to obtain \citep{masud2012facing, sethi2017reliable}. There have been proposed unsupervised drift detection methodologies \citep{spinosa2007olindda,masud2011classification,lee2012detection,ryu2012efficient,lindstrom2013drift,sethi2016grid}, which directly monitor the feature distribution of the data, to indicate drifts. These approaches suffer from excessive pessimism, as they cannot differentiate between drifts which affect the classifier's performance and those which do not. This problem of reliability of the drift detection, was addressed in the recent works of \citep{sethi2017reliable}, where the Margin Density Drift Detection (MD3) approach was developed. By including the context of classification, to the drift detection task, the MD3 approach was shown to provide domain independent and reliable indication of drift, from unlabeled data. Like most approaches in the area of concept drift detection \citep{goncalves2014comparative}, the domain agnostic nature of the MD3 approach was shown to be an advantage in \citep{sethi2017reliable}. However, in an adversarial environment, disregarding of domain characteristics, leads to missed opportunity and can make the detection process vulnerable to adversarial evasion. The drift is characterized by an attacker continuously trying to hide its trail, by learning about the behavior of the detection system first. As such, the drift detection can itself be vulnerable to adversarial manipulation at test time. Most drift detection methodologies are designed as adversarial agnostic approaches, where they consider drift to be independent of the deployed classifier. In an adversarial domain, the relation between the type of drift and the choice of the classifier $C$, is strongly coupled \citep{barreno2006can}. It is necessary to understand and incorporate the specific characteristics of adversarial drift, to make unsupervised drift detection applicable in such a domain.

An adversarial-aware, unsupervised drift detection approach, will take preemptive steps at the design of the classifier model, to ensure that future detection and retraining is made easier. In this paper, the \textit{Predict-Detect} framework is proposed, as an adversarial-aware unsupervised drift detection methodology, capable of signaling adversarial activity, with high reliability. The framework incorporates the ideas of reliability, which makes the MD3 \citep{sethi2017reliable} methodology effective, and presents a novel streaming data system, capable of providing life-long learning in adversarial environments. The developed methodology incorporates attack foresight into a preemptive design, to provide long term benefits for reactive security. The framework is developed as an ensemble based, application and classifier type independent approach, capable of working on streaming data with limited labeling. To the best of our knowledge, this is the first work which directly addresses the problem of adversarial concept drift in streaming data. The main contributions of the proposed work are as follows: 

\begin{itemize}
\item Analyzing the characteristics of adversarial concept drift, and the impact of the deployed classifier on the drifts generated at test time. 
\item Developing the \textit{Predict-Detect} classifier framework, as a novel approach to dealing with adversarial concept drift, in streaming data with limited labeling.
\item Extending the \textit{Predict-Detect} framework, to work with imbalanced data streams, by using active learning for labeling adversarial samples.
\item An novel simulation framework, for introducing systematic adversarial concept drift into real world datasets. This simulation framework allows for reusable testing, for better experimentation and analysis of the drift detection methodologies. 
\item Empirical evaluation of popular cybersecurity datasets, which provides avenues for further extension of the proposed methodology, to meet domain specific needs. 
\end{itemize}

The rest of the paper is organized as follows: Section~\ref{sec:lr} presents background work in the area of unsupervised concept drift detection and adversarial machine learning. In Section~\ref{sec:pm}, the proposed \textit{Predict-Detect} classifier framework is described. Experimental evaluation and results are presented in Section~\ref{sec:er}. Extension of the \textit{Predict-Detect} framework, to work with imbalanced adversarial drifting streams, is presented in Section~\ref{sec:al}. Additional discussion and ideas for extension are provided in Section~\ref{sec:discussion}, followed by conclusions in Section~\ref{sec:conclusion}.

\section{Background and related work}
\label{sec:lr}

In this section, the related research in the domain of concept drift and adversarial machine learning, is presented. Existing work on unsupervised concept drift detection, is presented in Section~\ref{sec:lr_cd}. The specific nature of adversarial activity, making it a special type of concept drift, is highlighted by recent works in the area of attacks on classifier systems, and is discussed in Section~\ref{sec:lr_adversarial}. Recent works incorporating adversarial awareness, in the design of dynamic systems, is presented in Section~\ref{sec:lr_acd}.

\subsection{Concept drift detection from unlabeled streaming data}
\label{sec:lr_cd}

Detection of concept drift is essential to dynamic classification systems, to provide high predictive performance throughout the progression of the data stream. Supervised classification techniques \citep{goncalves2014comparative}, are widely used in literature. These techniques rely on the continuous availability of labeled data, to compute performance (accuracy or f-measure), to monitor deviations from expected behavior. The over-dependence on labeled data, makes these techniques impractical for usage in a streaming data milieu, as labeling is an expensive and time consuming activity \citep{masud2012facing,lindstrom2010handling,kim2016concept}. To account for these limitations of supervised drift detection techniques, several unsupervised and semi-supervised drift detection methodologies have been proposed \citep{kim2016concept,sethi2016grid,masud2011classification,lee2012detection,ditzler2011hellinger,ryu2012efficient,lindstrom2013drift,qahtan2015pca}. These techniques rely on monitoring the distribution characteristics of the unlabeled data, to detect drifts. These are essentially change detection methodologies, which track deviations in the feature space distribution of the data, using statistical or clustering based techniques. As such, they are sensitive to any change in the data distribution, irrespective of its impact on the classification task. Not all changes in data result in a drop in the predictive performance of the classifier at test time, and as such the unsupervised techniques are prone to excessive paranoia which leads to false alarms. False alarms leads to wasted labeling effort, for validating drifts\citep{sethi2015don}, and also leads to loss of trust in the drift detection system. 

\begin{figure}[t]
  \centering
  \includegraphics[width=0.98\linewidth]{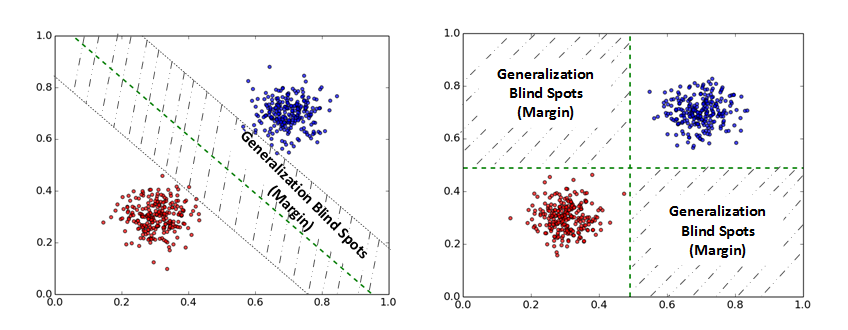}
   \caption{In MD3, the density of margin is tracked, to detect drifts from unlabeled data. \textit{left-} the margin region for SVM based classifier. \textit{right-} pseudo margin for classifier with no real notion of margin, defined as the disagreement region for a feature bagged ensemble.\citep{sethi2017reliable}}
  \label{fig:blindspots}
\end{figure}

To make unsupervised drift detection more reliable, the margin density drift detection algorithm (MD3) was proposed in \citep{sethi2017reliable}. The developed algorithm tracks the expected number of samples falling in the uncertain regions of a robust classifier (i.e. its margin), to detect changes which can affect the classifier's performance. For classifiers with explicit notion of margins, such as support vector machines (SVM), the margin is naturally defined and it's density can be tracked in a streaming environment \citep{sethi2015don}. For classifiers with no explicit notion of margin, such as C4.5 decision trees, the concept of a pseudo margin was introduced. By using a feature bagged ensemble, the margin was defined as the blindspot region, given by the region of high disagreement between the ensemble models (Figure~\ref{fig:blindspots}) \citep{tsethi}. Both the margin and the blindspot representation, were shown to provide similar benefits to unsupervised drift detection, making the MD3 methodology an effective classifier independent algorithm. Experimental evaluation demonstrated that the MD3 methodology is able to provide similar performance as a fully labeled approach (the Expert Weighted Moving Average approach EWMA \citep{ross2012exponentially}), and at the same time leads to lesser false alarm compared to unsupervised drift detectors (compared with the Hellinger Distance Drift Detection methodology \citep{ditzler2011hellinger}), making it more reliable and label efficient, for usage in a streaming data environment. 

The efficacy of the MD3 methodology, is a result of its ability to add the context of classification, to the task of drift detection. This is done by the inclusion of the margin characteristics, to discern if changes are relevant from the perspective of classifier performance. However, while the MD3 methodology includes the context of the learning task, it does not include any context about the nature of drifts that can occur at test time. This was seen as a strength of the methodology in \citep{sethi2017reliable}, making it more domain independent. However, in an adversarial domain, the drifts are more sophisticated, and can lead to evasion or misleading of the drift detection mechanism. Also, by not including the context of adversarial activity, the MD3 methodology misses out on preemptive design of classifiers, which can provide benefits post deployment.

\subsection{Adversarial manipulation of test time data}
\label{sec:lr_adversarial}

The security of machine learning has garnered recent interest from the research community, with several works demonstrating the vulnerabilities of classifiers to probing based attacks \citep{biggio2014pattern,tramer2016stealing,biggio2013evasion,papernot2016transferability,biggio2014security,tsethi2016}. These attacks are themselves data driven, and are caused by adversaries crafting samples at test time, which evade the deployed classifier of the defender, causing it to drop in predictive performance (accuracy). These attacks begin by the adversaries performing reconnaissance (exploration) on the classifier, to understand its behavior, and as such are referred to as exploratory attacks \citep{barreno2006can}. The information accumulated is then used to craft attack samples, leading to a drift at test time. These attacks are commonplace and pervasive, as they do not require any domain specific information or details about the internal workings of the defender's classifier model. The adversary, like any other client user of the system, accesses the system as black box system, to which it can submit samples and observe the response \citep{tsethi2016}. Such a setting was inspired from the design of cloud based machine learning services (such as Amazon AWS Machine Learning\footnote{\url{https://aws.amazon.com/machine-learning/}} and Google Cloud Platform\footnote{\url{cloud.google.com/machine-learning}}), which provide APIs for accessing predictive analytics as a service. These services were also seen to be vulnerable to exploratory attacks, where the adversary uses the client API to evade the black box model \citep{tramer2016stealing, sethi2017data}. 

\begin{figure}[t]
  \centering
  \includegraphics[width=0.95\linewidth]{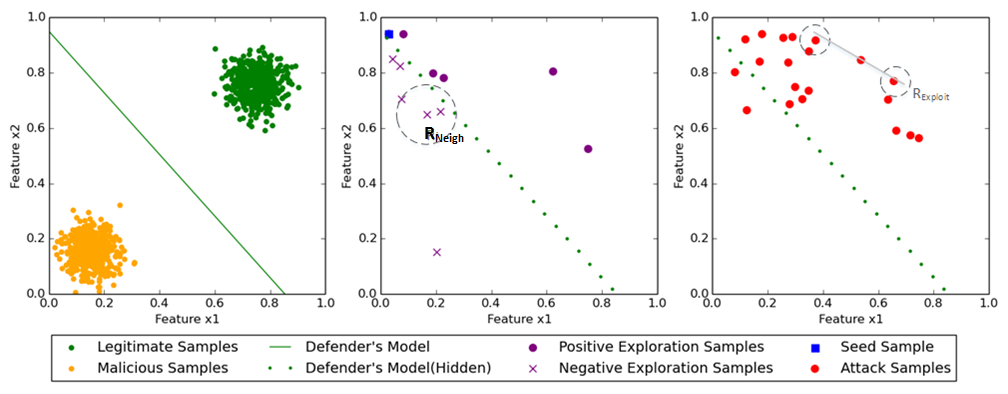}
   \caption{Illustration of AP attacks.\textit{(Left - Right)}: The defender's model from its training data. The Exploration phase depicting the seed (blue) and the anchor points samples (purple). The Exploitation attack phase samples (red) generated based on the anchor points.\citep{sethi2017data}}
  \label{fig:ap}
\end{figure}

The Anchor Points (AP) attacks of \citep{tsethi2016,sethi2017data}, is one such algorithm for simulating exploratory attacks on black box classifiers. The AP attacks uses an \textit{exploration-exploitation} framework, to perform evasion on classifiers, as shown in Figure~\ref{fig:ap}. The adversary starts with a seed sample, and then performs exploration by using a radius based incremental search approach. Each generated sample is submitted to the black box classifier of the defender, and its response is observed. Samples which are classified as \textit{Legitimate}, by the defender, are considered as ground truth values for the subsequent exploitation phase, and are referred to as the  Anchor Points. In the exploitation phase, the anchor points obtained in the exploration phase are used to generate additional attack points. The generation is based on a convex combination of the exploration points, and a perturbation operation, to add diversity to the final attack set of samples. The final set of attack points (red), as seen in Figure~\ref{fig:ap}, are submitted to the defender. These samples will cause the performance of the defender to drop, leading to a concept drift. 

The Anchor Points (AP) attacks, was developed as a methodology to analyze the vulnerability of machine learning systems, in \citep{sethi2017data}. It was presented from a static perspective, as batches of samples were submitted for exploration and exploitation. However, its analysis as a concept drift generation framework, has hitherto been ignored. Not only does the AP attack framework provide a way to generate adversarial concept drift, it also highlights an important characteristic of such drifts - the dependence of the generated drift on the initial trained classifier, which is explored and being evaded by the adversary. The extension of exploratory attacks to the domain of concept drift, has not been proposed elsewhere, to the best of our knowledge.

\subsection{Dealing with adversarial activity at test time}
\label{sec:lr_acd}

\begin{figure}[t]
\centering
\includegraphics[width=0.95\linewidth]{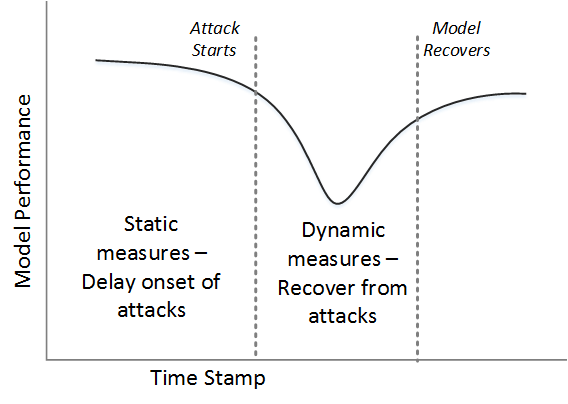}
\caption{Illustration of model performance over time, indicating onset of attack and recovery from it. Static measures of security focus on delaying onset of attacks. Dynamic measure focus on detection and recover only.}
\label{fig:static_dynamic}
\end{figure}

Adversarial activity causes the performance of the deployed classifier to degrade over time. Most works in adversarial machine learning, concentrate on making classifiers harder to evade. They do so by resorting to \textit{Complex learning} strategies \citep{papernot2016towards,rndic2014practical,wang2015robust,biggio2010multiple,biggio2010multipleattack, wozniak2014survey, stevens2013hardness}, which advocates integrating maximum informative features into the classifier models, or by integrating \textit{Randomness} into the prediction process \citep{biggio2014pattern,xu2014comparing, colbaugh2012predictability, vorobeychik2014optimal, papernot2016towards}. The emphasis of these two strategies, is to make attacks harder/expensive to carry out. However, these methodologies approach the problem of security from a static perspective. They focus on the ability to ward of attacks, but fail to provide insights or directions regarding measures to be taken after an attack commences. This is illustrated in Figure~\ref{fig:static_dynamic}, where the predictive performance of a classifier faced with an attack, and subsequent recovery, is shown. Static measures of \textit{Complex learning} and \textit{Randomness}, concentrate only on the portion of the attack-defense cycle before the attack starts. Any practical and usable system, also needs reactive dynamic measures, which can deal with attacks after it has affected the system's performance. Dynamic approaches are developed in the domain of concept drift research, where the aim is to detect and adapt to changes in the data distribution over time. One such methodology for reliable drift handling was developed in \citep{sethi2017reliable}, as the Margin Density Drift Detection framework (MD3).  However, this approach, like the other works on concept drift detection \citep{vzliobaite2010learning, gama2014survey, ditzler2015learning, goncalves2014comparative}, considers an adversarial agnostic view of the system. Concept drift detection considers any change in data equally, without regard into the domain specific nature of the change. This makes it ineffective in an adversarial setting, where drifts are a result of attacks, which are in turn a function of the model deployed by the defender. 

\begin{figure}[t]
  \centering
  \includegraphics[width=0.95\linewidth]{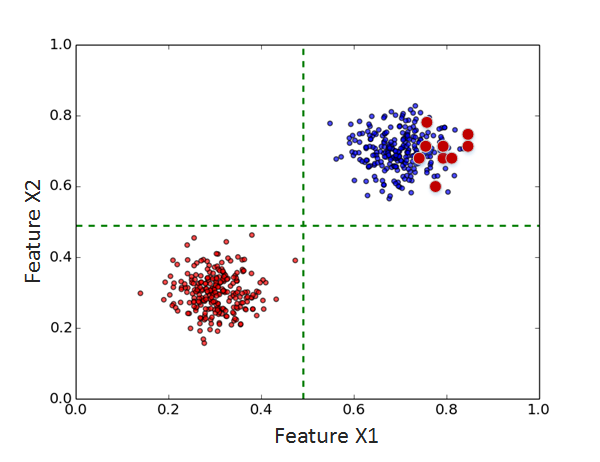}
   \caption{Illustration of \textit{Data Nullification} attacks. These attacks leads to corruption of the training data, and the inability to function in a streaming environment. \textit{blue-} legitimate training data, \textit{red-} malicious training data, and test time attack samples.}
  \label{fig:dn}
\end{figure}

Ideas and motivation regarding the dynamics of the machine learning security, were proposed in \citep{kantchelian2013approaches}. It was proposed that for a system to be of any practical use, it should evolve over time, and engage with human experts beyond feature engineering and labeling. A new taxonomy of exploratory attacks was presented in \citep{kantchelian2013approaches}, based on their effects on a dynamic data mining process. The idea of \textit{Data Nullification} attacks, was introduced. These attacks are a result of an adversary gaining excessive information about the training data samples, which it then uses to mimic benign samples perfectly. This causes attacks to overlap with the legitimate training data samples, leading to inseparability, and the inability of classifiers to distinguish between the two classes of samples (Figure~\ref{fig:dn}).  In such a case, 
the classifier cannot be retrained using the existing set of features, and new analysis and data collection is needed to redesign the system, which requires tedious examination of samples, to detect attacks and generate new set of features. From a dynamic machine learning perspective, the task of the defender should be to ensure that such type of attacks are avoided, even though evasion might be made slightly easier. This ensures that drifts will be detect-able and recover-able from. Works in the detection of adversarial activity were proposed in \citep{chinavle2009ensembles, kuncheva2008classifier, smutz2016tree}, which demonstrate that malicious activity can be detected by tracking disagreement scores from ensemble of models. However, these methods were not analyzed in a streaming environment, and also did not account for adversarial evasion capabilities at test time. 

The idea of data leakage and data nullification attacks, present a new dimension in the evaluation of machine learning security, beyond the traditional established metrics of hardness of evasion \citep{lowd2005adversarial}. These ideas are a result of a more widespread and practical understanding of the vulnerabilities of classifiers, resulting from black box and indiscriminate attacks at test time. Existing work on concept drift approaches this problem as a cyclic one, where data distribution changes are detected and handled over time \citep{vzliobaite2010learning}. However, concept drift research has hitherto taken a domain agnostic approach to dealing with distribution shifts. Adversarial characteristics of the drift, and preemptive strategies to mitigate and manage drift, have not been analyzed or discussed before, to the best of our knowledge. The only mention we found was in \citep{barreno2006can, kantchelian2013approaches}, where adversarial drift was understood to be a special type of drift, being a function of the deployed classifier itself. Since the attacks are targeted towards evading the deployed classifier, the choice of the deployed classifier directs the gamut of possible attack on the system, to a large extent. We analyze this specific aspect of the attacks, and design a preemptive strategy which benefits the dynamic handling of the attacks at test time.

\section{Proposed methodology}
\label{sec:pm}

In this section, the \textit{Predict-Detect} classifier framework is proposed. Section~\ref{sec:pm_motivation} presents motivation and intuition for the development of the framework. Development of the framework, as suitable for a streaming data environment, is presented in Section~\ref{sec:pm_framework}.

\subsection{Motivation - Misrepresenting feature importance}
\label{sec:pm_motivation}

\begin{figure}[t]
  \centering
  \includegraphics[width=0.95\linewidth]{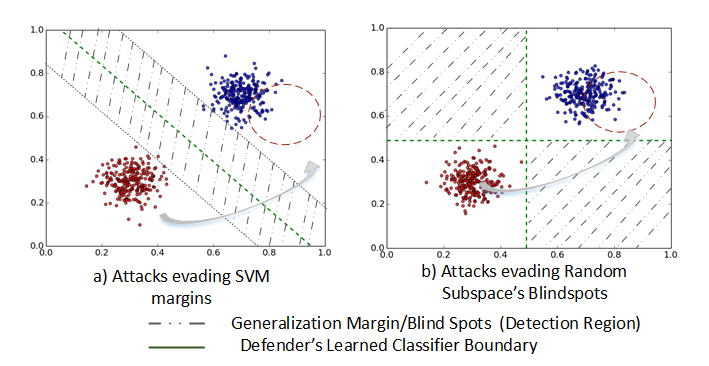}
   \caption{Drifts in adversarial environments will avoid low confidence regions, to avoid detection.}
  \label{fig:adverarial_drif_avoid}
\end{figure}

In the security of machine learning, robustness of classifiers is given by the effort needed by an adversary, to evade the model \citep{barreno2010security,lowd2005adversarial}. This has motivated the development and adoption of Complex Learning methodologies, which incorporate maximum feature information from the training data,  as a standard approach to securing classifier against probing based exploratory attacks. The intuition behind this strategy is as such: the inclusion of several informative features into the trained model makes the classifier more restrictive, as an adversary now has to simultaneously reverse engineer and mimic a majority of the benign data characteristics. While this intuition holds good from a static perspective, where we are trying to measure security in terms of the delay in the onset of attacks, it does not hold in a dynamic environment. In a dynamic environment, it is paramount for adversarial activity to be \textit{detect-able} and \textit{recover-able} from, so that a model is usable over time. Using the advocated Complex Learning strategies, excessive information is presented to the adversary, who over time can probe and learn the specific characteristics of the training data features. Such an adversary can accumulate the reconnaissance information, and then generate high confidence attacks, which avoid the uncertainty regions of the defender's model \citep{sethi2017reliable}. Such drifts are harder to detect and retrain from, as they overlap with the training - benign class data, and as such lead to a total compromise of the machine learning based system (Figure~\ref{fig:dn}). Figure~\ref{fig:adverarial_drif_avoid} illustrates one such adversary, which evades the MD3 approach of \citep{sethi2017reliable}. In MD3, it was proposed that margin density tracking over time can be used as a reliable indicator of concept drift from unlabeled data. This was shown to perform well over several concept drift detection tasks from different domains. However, the MD3 approach disregards the adversarial nature of drifts and as such could be fooled by an adversary, who avoids the uncertain regions of the deployed classifiers. 

The use of excessive information in the trained model, leads to leakage of information to the adversary, who uses carefully crafted probes to understand the behavior of the classifier model. For concept drift detection methodologies to function in adversarial domains, it is necessary for the test time adversarial samples to be separable from the benign training data samples. To ensure that the separability is maintained, it is necessary to limit the information available to be probed by an adversary. As an example of this requirement, consider the  following toy example: a 2-dimensional binary dataset, where the sample \textit{L(X1=1, X2=1)} represents the Legitimate class training sample and \textit{M(X1=0, X2=0)} represents the Malicious class sample. We evaluate two strategies for designing the defender's classifier $C$: a) A Complex Learning approach, which incorporates the information of both features \textit{X1} and \textit{X2}, into $C$, and b) A simpler model, which uses only the minimal amount of information necessary (i.e., one which uses either \textit{X1} or \textit{X2}, only). In case of the Complex Learning approach, the defender could either use a restrictive or a generalized strategy: $C:X1\cup X2$ or $C:X1\vee X2$. In either case, an adversary is able to understand the impact of the two features on $C$, and can generate an attack sample (1,1), to simultaneously evade both informative features. This attack sample is indistinguishable from the training data sample  $L$, causing unlabeled attack detection and handling, to fail. In the case of a simple model (here, $C:X1$), the attacker is able to evade the system by successfully mimicking $X1=1$. However, it is not completely certain about $L$, as no amount of probing will provide information about the specific impact of $X2$ (both (1,0) and (1,1) are accepted by $C$ as Legitimate). This example illustrates the advantages of a simpler model in adversarial domains,  as a counter intuitive but effective strategy to ensure dynamic functioning of classifiers.

\subsubsection{The \textit{Predict-Detect} classifier design}
\label{sec:pm_motivation_pd}

The nature of adversarial drifts makes it dependent on the characteristics of the deployed classifier model, which it is trying to evade \citep{barreno2006can}. As such, preemptive measures taken during the training of a classifier model, can benefit dynamic test time detection and handling of such drifts. The \textit{Predict-Detect} design uses this intuition, to develop classifier models which are able to detect adversarial activity at test time; reliably and with limited labeled data. By intentionally misrepresenting the un-informativeness of a subset of data features, this design is able to track adversarial activity.

\begin{figure}[t]
  \centering
  \includegraphics[width=0.95\linewidth]{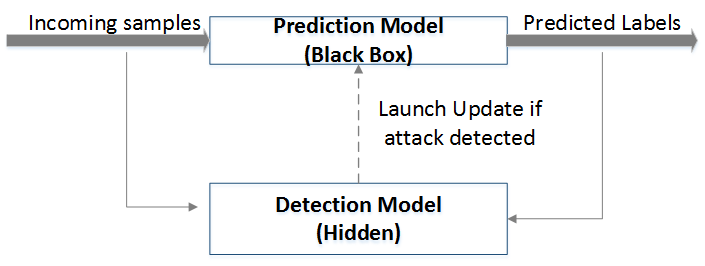}
   \caption{The \textit{Predict-Detect} classifier design. The \textit{Prediction Model} is deployed as the de-facto classifier of the system, and is prone to adversarial activity. The \textit{Detection Model} is used to track adversarial activity and detects when the system is compromised. }
  \label{fig:pd_usage}
\end{figure}

\begin{figure}[t]
  \centering
  \includegraphics[width=0.95\linewidth]{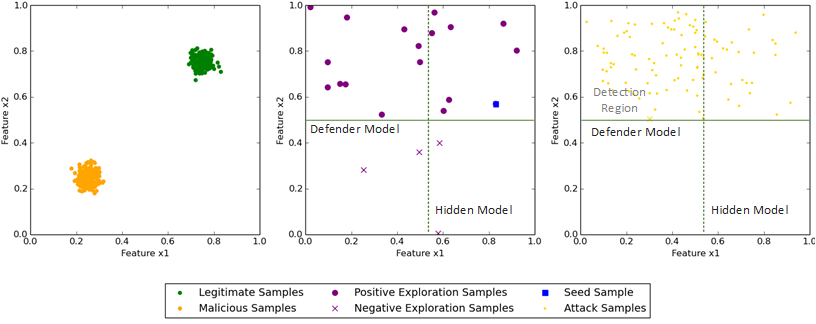}
   \caption{Illustration of the Predict-Detect classifier on 2D synthetic data. Significant portion of adversarial samples are captured by the \textit{Detection Region}.}
  \label{fig:pd_hidden}
\end{figure}

The \textit{Predict-Detect} design uses two orthogonal classifier, each trained on a disjoint subset of features of the training data, as shown in Figure~\ref{fig:pd_usage}. The first subset of features is used to form the defender's black box model, to perform prediction on the input samples. This classifier is called the \textit{Prediction} or the \textit{Defender} model, as its primary purpose is to perform prediction on the input samples, submitted by either the benign users or the adversary. Since this forms the black box model, it is vulnerable to probing based attacks by an adversary. The \textit{Prediction} model is expected to get attacked at some point, after deployment, due to the nature of the adversarial environment. The second subset of features is used to train a classifier, called the \textit{Detection} or the \textit{Hidden} model. This classifier is not used for any of the prediction task, and as such is shielded from external adversarial probes. This model represents information known by the defender, based on the original training data, but not accessible via probing to an adversary. The purpose of the \textit{Detection} model is to indicate adversarial activity, based on disagreement with the \textit{Prediction} model. Since an adversary launches an attack based on information learned from the black box, an attack is characterized by evasion of the \textit{Prediction} model, but only partial evasion of the \textit{Detection} model.  This is illustrated in Figure~\ref{fig:pd_hidden}, where a 2D synthetic training data is shown. Attacks on the black box \textit{Defender's} model are detected by an increase in the number of sample falling with the region of disagreement, called the \textit{Detection Region}. The two orthogonal models form a self monitoring scheme for detecting suspicious deviations in the data distribution, at test time.

It should be noted that this division of features in Figure~\ref{fig:pd_usage} is opaque to the adversary, who still submits probes on the entire feature set. The division of features is done internally by the framework. Thus no additional information is leaked to an adversary. Also, this framework does not advocate feature hiding, in accordance with the Kerchkoff's principles of information security \citep{kerckhoffs1883cryptographie,mrdovic2008kerckhoffs}, but instead relies on misrepresentation of the importance of the features, to the classification task. In case of Figure~\ref{fig:pd_hidden}, an adversary will know that the system consists of features \textit{X1} and \textit{X2}, but will not be able to ascertain the informativeness of feature \textit{X1}, no matter how long it probes the system. 

There proposed \textit{Predict-Detect} classifier design, aims to provide the following benefits: a) The misrepresentation of feature importance, will prevent drifts which overlap with the training data, preventing data nullification (Figure~\ref{fig:dn}), b) The uncertainty on the adversary's part will lead to samples falling in the \textit{Detection Region} at test time, enabling reliable unsupervised detection of adversarial drifts, c) In case of imbalanced data streams, where the attacks is a minority class, the \textit{Detection Region} provides a natural reserve for sampling the adversarial class data, enabling effective active learning over imbalanced data.

\subsection{The \textit{Predict-Detect} streaming classification framework for dynamic-adversarial environments}
\label{sec:pm_framework}

The proposed \textit{Predict-Detect} framework is developed as a streaming incremental framework for detecting and handling adversarial drifts. The overview of the framework is presented in Section~\ref{sec:pm_framework_overview}. Detailed design of the framework and its major components is presented in Sections~\ref{sec:pm_framework_generate}, \ref{sec:pm_framework_detect} and \ref{sec:pm_framework_retrain}.

\subsubsection{The \textit{Predict-Detect} streaming classification framework}
\label{sec:pm_framework_overview}

\begin{figure}[t]
  \centering
  \includegraphics[width=0.95\linewidth]{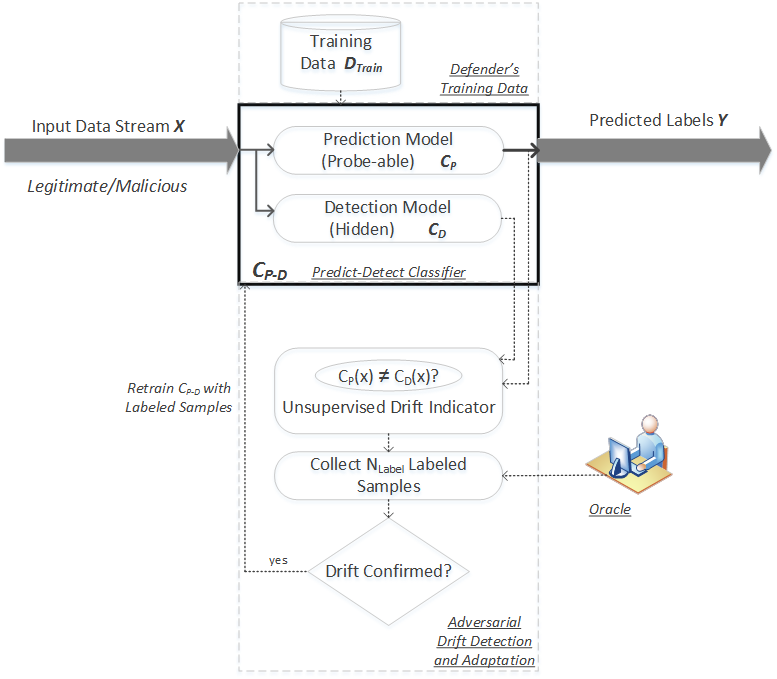}
   \caption{Overview of the \textit{Predict-Detect} stream classification framework.}
  \label{fig:pd_overview}
\end{figure}

The overview of the \textit{Predict-Detect} streaming framework is presented in Figure~\ref{fig:pd_overview}. The input stream of unlabeled data samples $X$ is processed, leading to the output predicted label stream $Y$. In this setup, the framework has an unsupervised drift indication component, which processes $X$ to detect if there is any significant drifts, which could cause the predictive performance to drop. Upon signaling a drift, the framework requests additional labels ($N_{Labels}$), from an external oracle, often at a price (cost and time resources). These labeled samples are used for confirmation of the earlier signaled drift (by the unsupervised component). If a significant drop in the predictive performance (accuracy/f-measure)  is seen over the labeled samples, the framework confirms the occurrence of an adversarial drift. In the event of a confirmed drift, the labeled samples are used to retrain the predictive model, to ensure the continued efficacy of predictive performance of the system. Since the labeled samples are requested only when a drift is first suspected, using unlabeled data, this framework prevents wastage of labeling effort, which results from continuous monitoring of the stream \citep{sethi2015don}. In the event of infrequent drifts and long periods of stationarity, the framework will not indicate drifts, and thereby no labeling will be wasted. 

The core design aspect of the framework, which makes it effective in an adversarial environment, is the development of the \textit{Predict-Detect} classifier model, as seen in Figure~\ref{fig:pd_overview} as the $C_{P-D}$ model. The framework leverages the benefits of the \textit{Predict-Detect} design of Section~\ref{sec:pm_motivation_pd}, as a novel mechanism to hide feature importance information from the adversary, and extends it to be applicable in a streaming data domain. The classifier relies on a coupled model strategy, where two orthogonal models are trained from the training data $D_{Train}$. The first model is called the \textit{Prediction} model ($C_P$), and it is trained on one disjoint subset of features of the original training dataset. The other subset of features, not used in ($C_P$), is used to train the \textit{Detection} model ($C_D$). The \textit{Prediction} model is the defacto model of the framework, used for classifying the input unlabeled stream $X$, to produce output labels $Y$. Since this is the prediction model facing the input stream of data, this model is susceptible to adversarial activity, at test time. This forward facing model is expected to be attacked, by an adversary using probing based exploratory attacks, and observing the feedback provided by the framework. As such, the model's performance is expected to drop over time. The \textit{Detection} model is shielded from the probes of the adversary, as it is kept hidden and away from the prediction process. As seen in  Figure~\ref{fig:pd_overview}, the feedback of the model $C_{D}$ is not presented to the outside world. This model is used for the detection of adversarial activity. An adversary, using probes to understand the behavior of the black box model $C_P$, will fail to successfully reverse engineer $C_D$, which is trained on an orthogonal subset of features. As such, the adversary will not be able to exactly mimic the training data characteristics. This will enable the tracking of adversarial activity, indicated as an increase in the disagreement between the predictions of the two models. Input data stream $X$ is split between the two models, vertically based on the features used to train the respective model, and the prediction by $C_P$, is presented as the output of the framework. No amount of probing will enable the adversary to understand the significance of features held by $C_D$, as they are not a part of the prediction process of the framework. The misrepresentation of these feature importance, in an adversarial domain, is the core of what makes this framework effective. 

\subsubsection{Generating the \textit{Prediction} and the \textit{Detection} models from the training data}
\label{sec:pm_framework_generate}

The training dataset, containing $F$ features, is divided into two subsets (vertically based on features), to train the \textit{Prediction} and the \textit{Detection} models. Such splits of training data are possible on most cybersecurity datasets, due to the presence of multi-modal and orthogonal informative features, in high dimensional datasets. We intend to define a division of the feature space, such that each of the trained models has high predictive performance. Such divisions are naturally defined when the training dataset is aggregated from multiple sources, with clearly defined boundaries. An example of such a system is a multi-modal biometric system \citep{joshi2009multimodal}, which uses both face recognition and fingerprint scanners to provide the final authentication. In such a system, it is easy to split the features into two disjoint subsets, one for the face recognition, and the other for fingerprint data. 

\begin{algorithm}[t]
\SetKwInOut{Input}{Input}
\SetKwInOut{Output}{Output}
 \Input{Training data ($D_{Train}$) with features $F (1..k)$, Number of splits $N$ (For the $C_{P-D}$ classifier, $N$=2.)  }
 \Output{Trained models $M_{Trained}$}
 
 $F_{Splits}$ = [[]] 
 
 $M_{Trained}$ = []
	
 $F_{sort\_importance}$ = \textit{RankByFeatureImportance}($F$)
 
 \For{ i = 1 .. $N$}{
	  $F_{Splits}$[i\%N].add($F_{sort\_importance}$[i])\\
	  \Comment{Round robin split of ranked features}
 }

 \For{ i = 1 .. $N$}{
 	  $D_{Train\_split}$ = $D_{Train}$
 	  
 	\For{feature in $F$}{
 		\If {feature not in $F_{Splits}$[i] }{
 			Blank out $D_{Train\_split}$, by replacing with default value \\
 			\Comment{\textit{`blanking-out'} non associated features}
 		}
    }
    $M_{Trained}[i]$ $\leftarrow$ Train model on $D_{Train\_split}$
 }  
 \Return $M_{Trained}$
\caption {Generating multiple models from training data, by splitting features. }
\label{algo:generate_pd}
\end{algorithm}

In the absence of domain specific knowledge of the features, random partitioning is usually resorted to. However, this is not optimal, as a majority of the informative features can end up clumped together in the same partition. We propose a feature ranking based approach, which considers dividing the features uniformly based on their importance to the classification task. We use feature ranking to rank the initial set of $F$ features, based on the training data. We then distribute the features in a round robin fashion, to form the two subsets: $F_P$, used to train the \textit{Prediction} model, and $F_D$, for training the \textit{Detection} model. While several feature ranking approaches are available, the F-value from ANOVA is considered here for experimentation purposes, as the methodology for measuring feature importance\footnote{Using scikit-learn's \citep{scikit-learn} \textit{sklearn.feature\_selection.SelectKBest} function, to score all features using \textit{sklearn.feature\_selection.f\_classif}.}, without loss of generality. 

\begin{figure}[t]
  \centering
  \includegraphics[width=0.95\linewidth]{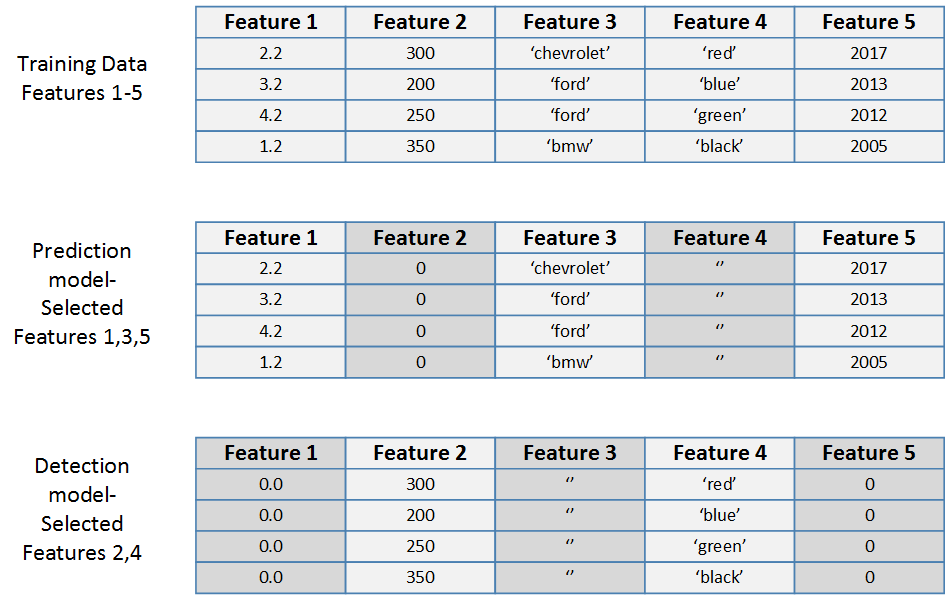}
   \caption{Illustration of feature splitting between the \textit{Prediction} and the \textit{Detection} model. The blanked out features of each model are highlighted. }
  \label{fig:pm_generate}
\end{figure}

The splitting of features and generation of orthogonal trained models, is done according to Algorithm~\ref{algo:generate_pd}. The set of $F$ features $(1..k)$, is divided to form multiple disjoint subsets, based on their informativeness to the prediction task, obtained from the training data. For the \textit{Predict-Detect} classifier, two subsets are needed: $F_P$ for the \textit{Prediction} model ($C_P$), and  $F_D$ for the \textit{Detection} model ($C_D$) ($F=F_P \cup F_D$). As such each model has an associated subset of features, which it deems important. The original training data is modified, such that only the associated features are used in training a model. This is done by the \textit{`blank-out'} process, given in Line 11. This process takes the original training dataset, and replaces all non model associated features with a predefined default value. An example of this process is depicted in Figure~\ref{fig:pm_generate}. The original set of 5 features is divided into 2 disjoint subsets. The \textit{Prediction} model is associated with Features 1,3 and 5 ($F_P=\{1,3,5\}$). As such, these features are retained in the training data, and the remaining are filled with default values, for all data samples. This process nullifies the discriminatory information of the Features 2 and 4, thereby barring them from being included in the \textit{Prediction} model. Similarly, for the \textit{Detection} model, Features 2 and 4 are included in the model ($F_D=\{2,4\}$), while Features 1,3 and 5 are blanked out. Once the two models are trained on the modified training data, they can each receive unlabeled stream samples of $\left| F  \right| $ features. The ability to disregard features is built into the training of the two models. The advantage of this \textit{`blank-out'} process, as opposed to splitting each incoming sample into two subsets based on features, is that now each of these models operate on the entire feature space $F$ and can receive the same input sample $X$. The training of the models has equipped it with the ability to assign importance to their required feature set only. The Algorithm~\ref{algo:generate_pd} is general in its presentation, as it allows the training of $N$ disjoint models. For the \textit{Predict-Detect} classifier, we take $N$=2, and use one of the trained models for prediction, while the other for detection.

In the absence of domain specific knowledge and correlation information, we can ensure that important features are evenly distributed among the two sets of models, using Algorithm~\ref{algo:generate_pd}. In real world systems, where correlation between features can cause attributes to change in tandem, more sophisticated feature splitting techniques can be employed. The feature subset ensemble techniques of  \citep{kons2014enhancing} combined with the cluster based feature splitting of \citep{aly2006novel}, could be used to form multiple classifiers with uncorrelated features within them. For the effective usage of the \textit{Predict-Detect} framework, we only need to ensure that the generated splits of features are disjoint and result in good predictive performance for each of the models.

\subsubsection{Detecting adversarial drift reliably from unlabeled data}
\label{sec:pm_framework_detect}

The proposed framework relies on detection of drifts from the unlabeled data stream, to save expenditure of labeling budget on validation of the prediction model. This is made possible by the coupled setup of the \textit{Prediction} and the \textit{Detection} models. An increased disagreement between the predictions of the two models, on new incoming samples, is suspicious and indicates a possible drift. The tracking of this disagreement, over time, in a streaming environment is used for unsupervised adversarial drift detection. This is presented in Algorithm~\ref{algo:ad_detect}, which is motivated in setup by the margin density drift detection algorithm (MD3) of \citep{sethi2017reliable}. The MD3 algorithm tracks the number of samples falling in a robust classifier's margin, in a streaming environment, to indicate the possibility of a drift. The \textit{Predict-Detect} framework provides an adversarial aware drift detection mechanism over unsupervised streams, by causing attacks to be detected based on disagreement with the hidden\textit{Detection} classifier model.

\begin{algorithm*}[t]
\SetKwInOut{Input}{Input}
\SetKwInOut{Output}{Output}
 \Input{ Unlabeled stream X, Predict - Detect models $C_P, C_D$, Reference distribution ($PD_{Ref}$, $\sigma_{Ref}$, $Acc_{Ref}$, $\sigma_{Acc}$). \textit{Parameters:} Sensitivity $\theta$, Stream progression $\lambda=(N-1)/N$ (where N is the chunk size), $N_{train}$ (= N by default), $N_{unlabeled}$ (= N by default)}
 \Output{Predicted label stream Y}
 
 $PD_0$ = $PD_{Ref}$
 
 currently\_ drifting = False
  
 \For{ t= 1,2,3,...:}{
 
 	Compute disagreement score - 
 	$	Dis(x=X_{ t })=\begin{cases} 1,\quad if\quad C_P(x)!=C_D(x) \\ 0,\quad otherwise \end{cases}
	$
		
	Update $PD_t$ = $\lambda$ * $PD_{t-1}+(1-\lambda)*Dis(X_t)$
	
	\If{$|PD_t-PD_{Ref}|\textgreater \theta$ * $\sigma_{Ref}$ and not currently\_drifting}
	{
		currently\_drifting = True \\ \Comment{\textbf{Drift Suspected}}
		
		Collected\_unlabeled\_samples=0
		
		$D_{Unlabeled}$ = $\emptyset$
		
		$D_{Labeled}$ = $\emptyset$
		
	}
	\If{currently\_ drifting and Collected\_unlabeled\_samples  $<N_{unlabeled}$}
	{
		$D_{Unlabeled} \quad \cup \quad X_t$ \\ \Comment{\textbf{Collect samples to be labeled}}
		
		Collected\_unlabeled\_samples ++
		
	}
	\ElseIf{currently\_drifting}
	{	\Comment{Enough samples to make decision}

		$D_{Labeled} \quad \leftarrow$ Label samples from $D_{Unlabeled}$, using \textit{Oracle}, upto $N_{train}$	
			\\
			\Comment{\textbf{Active learning} can be used in case $|D_{Unlabeled}|>N_{train}$ }
		
		\If{($Acc_{Ref}$-$Acc_{D_{Labeled}}$)$\textgreater \theta$  * $\sigma_{Acc}$}
		{
			Retrain $C_P, C_D$ with $D_{Labeled}$ 
			\\
			\Comment{\textbf{Drift Confirmed}}
		}
		Update Reference distribution ($MD_{Ref}$, $\sigma_{Ref}$, $Acc_{Ref}$, $\sigma_{Ref}$)
		
		currently\_ drifting = False		
	}
	}
	\Return $C_P(x)$
\caption {Unsupervised drift detection in the \textit{Predict-Detect} framework.}
\label{algo:ad_detect}
\end{algorithm*}

The unsupervised drift detection mechanism is presented in Algorithm~\ref{algo:ad_detect}. The \textit{Prediction} and the \textit{Detection} models ($C_P$ and $C_D$), generated from the initial training data, are used for the detection of drifts from the unlabeled data stream $X$. Also, the training data is used to learn the expected disagreement and acceptable deviation ($PD_{Ref}$ and $\sigma_{Ref}$), along with the expected prediction performance (measured in accuracy for balanced streams), given by $Acc_{Ref}$ and $\sigma_{Acc}$. This information is learned via 10-fold cross validation, and is used to characterize the normal behavior of the stream. This is done by dividing the training data into 10 bands, and generating the $C_P$ and $C_D$ models on 9 bands at a time, and then computing the disagreement on the 10th band. This value is obtained 10 times, and the reference distribution is learned from it. This serves as a basis for establishing normal expected behavior, from which drift can be detected based on deviation. Sensitivity is controlled using the parameter $\theta$, which provides a user friendly way to specify acceptable deviation of the stream, in terms of reference characteristics learned from the initial training data. As such, the framework allows specifying data independent parameters, making it more intuitive to the end user. 

For each incoming samples $x$, the disagreement in predictions between the $C_P$ and the $C_D$, is computed as the signal $Dis$. This disagreement is aggregated over time, to  form the $PD_t$ metric, as computed in Line 5. The computation uses a time decaying incremental tracking of the metric PD, dictated by the chunk size $N$. A sudden increase in the disagreement metric $PD$, is indicative of an adversarial drift. This indication is controlled by the sensitivity parameter $\theta$, specified by the user based on its tolerance for deviation. The initial indication of drift given by Line 6, is the unsupervised drift indicator component of the framework. It goads the system administrator to examine the data for potential attacks. Attacks are confirmed by collecting $N_{Unlabeled}$ samples and labeling them to form the labeled dataset ($D_{Labeled}$).  This $D_{Labeled}$ set of labeled samples is used to confirm drifts, and to retrain the models of the framework in case a drift is confirmed. A drift is confirmed if the predictive performance (measured as accuracy here) on the labeled samples is seen to fall significantly. This fall  is measured in terms of deviation from the reference accuracy ($Acc_{Ref}$), which was obtained from the training data. 

Once a drift is confirmed, by testing the predictive performance on the obtained labeled data, the models (i.e., the \textit{Prediction} and the\textit{Detection} models) need to be retrained, to represent the new distribution of the stream. Relearning is performed using the obtained labeled samples $D_{Labeled}$, by querying the Oracle. Querying the Oracle is expensive as it requires time/effort to obtain expert feedback on the unlabeled samples. Since the proposed algorithm request labels only for drift confirmation and relearning, it provides a cost effective and practical methodology for application in streaming data domains, as it does not waste any labeling for constantly checking for drifts in every time period. In the Algorithm~\ref{algo:ad_detect}, the labeling of samples follows a naive strategy, where the $N$ subsequent samples after a drift indication are requested to be labeled by the Oracle, to form the labeled set $D_{Labeled}$. The modular design of the framework leaves it open to extension with other active learning techniques, which can further dictate how the samples are labeled after a drift is indicated.
 
The entire drift handling process is kept internal to the black box system, and the end user/adversary is agnostic to the adaptive mechanism of the framework. The end user/adversary is provided prediction on the input samples $x$, using the prediction model $C_P(x)$. Since additional labels are only requested when an attack is suspected, the framework works in an unsupervised manner for the majority of the stream, without the need for constant labeled validation. Only when drifts are suspected and retraining may be needed, are labeled samples requested. This makes the framework attractive for usage in dynamic adversarial drifting environments, where labeling is expensive and time taking. 

\subsubsection{Retraining the \textit{Predict-Detect} models - Drift Recovery}
\label{sec:pm_framework_retrain}

In order to recover from the effects of an adversarial drift, the classifier models of the framework need to be retrained, using the obtained labeled data. The following strategies and their impact on the dynamic learning process are discussed, as potential retraining options.

\begin{itemize}
\item \textit{Using the existing feature split (\textbf{PD-NoShuffle})}: In this strategy, the existing feature splits of the \textit{Prediction} and the \textit{Detection} models, are retained. The models are trained on the new labeled data, based on the same set of features already assigned to them. This strategy has the advantage to keep features in the \textit{Detection} model hidden from an adversary, throughout the progression of the stream. However, this strategy does not account for the changes in the feature ranking and can lead to poorly trained classifier models, which are a result of changes in the importance of features over time, following drifts in data. 
\item \textit{Re-splitting features (\textbf{PD-Shuffle})}: Here, the framework uses the labeled data to regenerate feature splits based on Algorithm~\ref{algo:generate_pd}. The newly split models are then used for deploying the \textit{Prediction} and the \textit{Detection} models. This strategy assumes every drift recovery to be an independent start of a new attack-defense cycle. This is especially true when a system is faced with multiple independent adversaries, over time. This strategy is assumed to provide better trained models than the \textit{PD-NoShuffle}, as it has the opportunity to re-calibrate feature ranking and generate new splits of the features, after every attack cycle. 

\end{itemize}

While both strategies have their shortcomings and advantages, a defender can also resort to a combined approach while applying this framework.  First the defender can check to see if it is able to use the existing feature splits to retrain the model and receive sufficient predictive performance, using cross validation on the training dataset. If the features are rendered unusable due to a sophisticated attack, the features can be re-split. Delaying the shuffling of features is advantageous, as shuffling transfers information about the hidden features onto the prediction forefront. An adversary could use this information over time to generate a potent temporal attack, by aggregating information from multiple cycles of attacks. While we have not found any documentation for such a category of attacks, it is a possibility, and as such we should avoid shuffling of features whenever possible. 

In either case, the adversary is devoid of complete information, which makes it unable to launch a data nullification attack \citep{kantchelian2013approaches}. This ensures that adversarial activity will be detected and that retraining will be possible, for continued operation in streaming domains. By constantly gaming the adversary and responding quickly to attacks, the attacks are rendered expensive and futile, making this reactive system an attractive defense strategy for securing against adversarial data drifts. 

\section{Experimental evaluation}
\label{sec:er}

Experimental evaluation and comparison of the proposed framework, is presented in this section. Section~\ref{sec:er_generate} presents a framework for simulating adversarial concept drift in real world datasets. Setup and methods used for comparative analysis, is presented in Section~\ref{sec:er_analysis}. Results on data streams with single adversarial drift is presented in Section~\ref{sec:er_single}, and those with multiple adversarial drifts is presented in Section~\ref{sec:er_multiple}.

\subsection{Generating adversarial concept drifts on real world datasets}
\label{sec:er_generate}

Adversarial concept drift is a special type of concept drift, as the data distribution changes are introduced by an adversary aiming to subvert the performance of the deployed classifier. As such, these distribution changes are dependent on the classifier trained and deployed by the defender. An adversary begins the attack cycle by probing the deployed black box model of the defender, and then uses this information to generate samples which evade detection. This characteristic of adversarial drift makes it a special category of concept drifts, which needs to be analyzed and dealt with differently, than regular concept drift. It is not possible to evaluate these drifts on existing datasets (which are popularly used for concept drift research), as adversarial drifts are dependent on the deployed classifier model, which they are trying to evade. As such, we present here, a strategy for simulating adversarial drifts on real world datasets. We do this by extending the Anchor Points (AP) attack framework of \citep{tsethi2016}. The AP framework was developed for simulating data driven exploratory attacks, on black box classifier models. The framework was developed for static evaluation, as the attack samples were only evaluated for the purposes of demonstrating the possibility of classifier evasion, by an adversary. We extend the AP framework, to be used in a streaming environment, under the \textit{AP-Stream} framework, presented here. 

\begin{figure*}[t]
  \centering
  \includegraphics[width=0.85\linewidth]{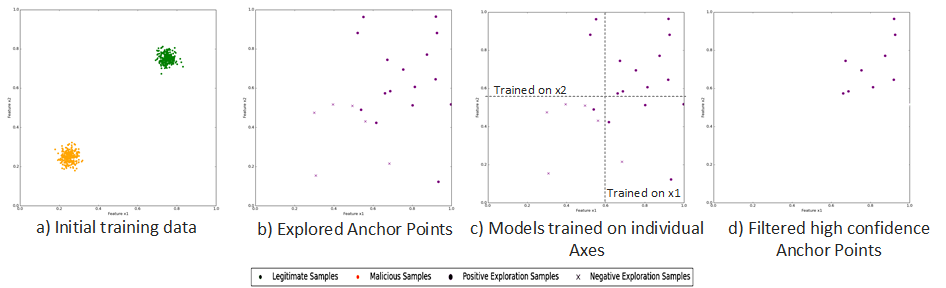}
   \caption{Reducing adversarial uncertainty via filtering, in Anchor Points attacks \citep{tsethi2016}. After the exploration phase in b), the adversary trains models on individual feature subspaces (X1 and X2). It then aggregates this information to clean out samples in low confidence areas (X1$\ne$ X2). The final set of filtered high confidence samples are shown in d).}
  \label{fig:er_filter}
\end{figure*}

There are two considerations in the extension of the Anchor Points attacks framework for the simulation of adversarial concept drift: a) Accounting for sophisticated adversarial activity, aiming to evade drift detection by uncertainty tracking unsupervised approaches, and b) Converting the static attack framework to a streaming data representation. For the former, we take into account adversaries who first perform exploration of the black box classifier, and then filters out low confidence attack samples. The confidence is determined on the adversary's side, by training a random subspace ensemble on the exploration samples, and then filtering out samples of high disagreement between the component models of the ensemble. This is demonstrated in Figure~\ref{fig:er_filter}. The adversary first uses the Anchor Points (AP) attack framework to generate the exploration samples of Figure~\ref{fig:er_filter} b). It then proceeds to train a feature bagged (random subspace) ensemble model on the samples in b), and then eliminates samples of high disagreement. The resulting high confidence exploration samples are shown in Figure~\ref{fig:er_filter} d). These exploration samples represent the most confident and best reverse engineered samples by the adversary, and it uses these samples to exploit and generate the attack campaign. 

\begin{figure}[t]
  \centering
  \includegraphics[width=0.95\linewidth]{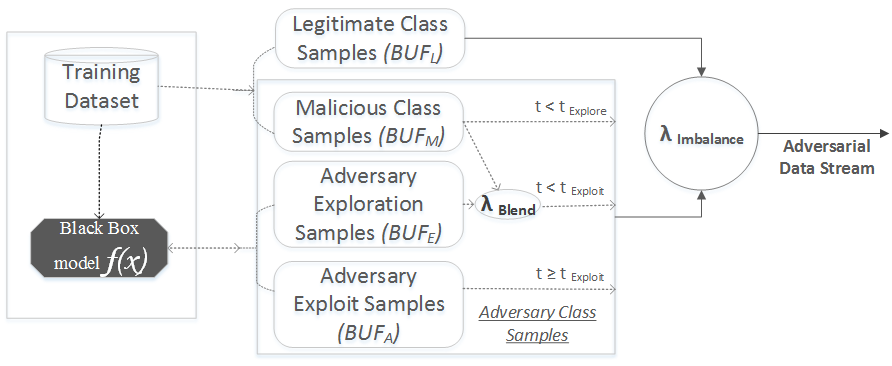}
   \caption{The \textit{AP-Stream} framework for simulating adversarial concept drift.}
  \label{fig:er_simulate}
\end{figure}

To extend the Anchor Points (AP) attacks to a streaming environment, the \textit{AP-Stream} framework is developed as a wrapper over the already developed AP framework.  The \textit{AP-Stream} framework receives samples from the attack simulation on the defender's black box, and it converts these samples into a stream of data, for adversarial analysis, as shown  in Figure~\ref{fig:er_simulate}.  The initial training data, from the real world dataset, is split into two parts: the \textit{Legitimate} class samples and the \textit{Malicious} class samples. These splits are used to form the initial distribution of the stream, before the drift starts. The training dataset is also used to train the defender's model ($f(x)$). The defender's model is then attacked using the AP framework, and the resulting exploration and exploitation samples are stored in the corresponding buffers ($BUF_E$ and $BUF_A$, respectively), as shown in Figure~\ref{fig:er_simulate}. 

Data stream samples are generated by random sampling from the 4 different data buffers of Figure~\ref{fig:er_simulate}. At any given time, the legitimate class samples are obtained from sampling the buffer $BUF_L$, and the adversary's samples are sampled from one of $BUF_M$, $BUF_{E}$ or $BUF_{A}$. The parameter $\lambda_{Imbalance}$, is used to control the amount of imbalance between the legitimate and adversarial class samples, in the stream. Since, we assume that the legitimate class does not drift over the course of the stream, we draw the regular class samples by always sampling from the initial pool of legitimate class data ($BUF_L$) obtained from the training dataset. However, for the adversarial class samples, the samples are drawn from the three different buffers, over the course of the stream. Initially, till time $t_{Explore}$, samples are drawn from the original pool of the malicious class training samples ($BUF_M$). This is the time period where an adversary has not yet started the attack process. After $t_{Explore}$, we draw samples from both the Malicious class training samples ($BUF_M$) and the pool of exploration samples ($BUF_E$). The rate at which the samples are drawn from each of these buffers is controlled by the parameter $\lambda_{Blend}$, called the blending rate. By blending exploration samples with the original set of malicious samples, an adversary avoids detection in the exploration phase. After the adversary obtains enough information about the black box model (upto the exploration budget $B_{Explore}$), it starts the exploitation phase, which is the attack payload for this adversarial cycle. This is done after time $t_{Exploit}$,  by sampling from the pool of exploitation samples ($BUF_A$). The time values $t_{Exploit}$ and $t_{Explore}$, enable the user to set up and control the profile of the stream, that they want to simulate. 

\begin{figure}[t]
  \centering
  \includegraphics[width=0.95\linewidth]{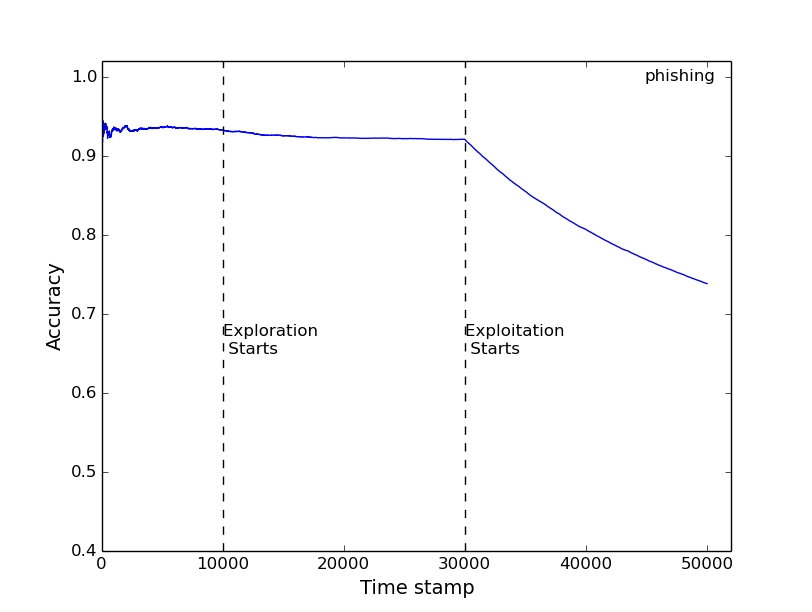}
   \caption{Simulation of adversarial concept drift on the \textit{phishing} dataset \citep{Lichman:2013}.}
  \label{fig:er_simulate_phishing}
\end{figure}

The simulation of adversarial drift on the \textit{phishing} dataset \citep{Lichman:2013}, is shown in Figure~\ref{fig:er_simulate_phishing}. The stream demonstrates an exploration phase till $t_{Explore}$=10,000 samples, and the attack phase starts at $t_{Exploit}$=30,000 samples. The data is taken to be balanced ($\lambda_{Imbalance}=0.5$) and the blend rate is taken to be $\lambda_{Blend}=0.05$ (i.e., 5\% of the sample in the exploration phase are drawn from the exploration buffer). From the figure, the effects of adversarial drift can be seen starting at $t$=30,000, as the accuracy starts to rapidly drop. This is a result of the adversarial manipulation of samples, to evade the deployed black box model. In order for a model to be usable, it is necessary to detect and fix the effects of such adversarial drifts. The drop in accuracy at the exploration phase is minimal, and can go undetected, as is intended by an adversary. In case of an overly sensitive detection system, the adversary over time can learn to use a lower blend ratio, to cover its tracks. 

Using the proposed \textit{AP-Stream} framework,  we can simulate the AP framework to work in a temporal environment, and can adjust the characteristics of the stream, to analyze particular aspects of different security measures. 

\subsection{Experimental methods and setup}
\label{sec:er_analysis}

This section presents the experimental methods and parameters, used for the analysis of the proposed framework. Methodologies used for comparative analysis are presented in Section~\ref{sec:er_methods}. Description of datasets and experimental setup, are presented in Section~\ref{sec:er_setup}.

\subsubsection{Methodologies used for comparative analysis}
\label{sec:er_methods}
~

The effects of adversarial drifts, on the classification performance over time, is demonstrated by experimentally comparing the following drift handling methodologies.  

\begin{itemize}
\item \textit{Static baseline model (\textbf{NoChange}):} This methodology is overly optimistic, as it assumes that the data will never drift ,and that the initial trained model is sufficient to retain performance over time. This is an unrealistic assumption, but this serves as a lower baseline for evaluating other drift handling strategies. Any proposed methodology should be atleast as good as this strategy, to be of any real use.
\item \textit{Fully labeled accuracy tracking (\textbf{AccTr}):} This serves as an upper baseline for our evaluations, as it represents an optimal case, where all the data is labeled, and the labels are immediately available after the prediction is made on an input sample. This model tracks the classifier's predictive performance (e.g., Accuracy), to signal drift. An unsupervised technique is considered effective, if it provides performance close to the AccTr approach, while reducing labeling requirements. For this methodology, the accuracy is tracked incrementally by using the EWMA \citep{ross2012exponentially} formulation of change tracking.
\item \textit{Margin density drift detection (\textbf{MD3-RS})}: The MD3 methodology was proposed in \citep{sethi2017reliable}, as an alternate to traditional unsupervised drift detectors. MD3 was shown to be more reliable than traditional distribution tracking methodologies, as it tacitly involves the classifier's notion of uncertainty into the drift detection process. The MD3-RS methodology uses a random subspace ensemble for the detection purposes. A sudden increase in the number of samples in the ensemble's margin (i.e., region of disagreement), is considered to be indicative of a drift. Comparing with MD3 provides us with insights into the need for adversarial awareness in drift detection. In our experiments, we consider a random subspace ensemble of 50 linear SVMs, with 50\% of the features in each base model. The threshold for the certainty margin is taken to be as 0.5.
\item \textit{The Predict-Detect framework without feature shuffling for retraining (\textbf{PD-NoShuffle}):}  This is the proposed \textit{Predict-Detect} classifier framework, with 50\% of the features belonging to the \textit{Prediction model}, and the other 50\% belonging to the hidden \textit{Detection model}. The drift is detected based on tracking the disagreement between the two models. Upon drift confirmation, the individual models are retrained without regenerating the feature splits. This methodology evaluates the ability and effects of continuing to use the same set of initially split features, so as to ensure feature importance hiding, for a longer period of time. 
\item \textit{The Predict-Detect framework with feature shuffling for retraining (\textbf{PD-Shuffle}):} This is the proposed framework similar to PD-NoShuffle, except for the fact that retraining involves reshuffling all features, and then regenerating the two separate prediction and detection models. This model evaluates the impact of ignoring temporal information gained by an adversary, and focuses on impact against multiple independent adversaries over time. 
\end{itemize}

The \textit{Prediction} and the \textit{Detection} model in the proposed methodology, are comprised of a random subspace ensemble of 50 linear SVMs (L1 penalty, regularization constant c=1), each with 50\% of the features randomly allocated to them. We use the same ensemble framework for the MD3 classification model, as well as for the prediction model for the AccTr and the NoChange model, to ensure consistency in evaluating the methods. 

\subsubsection{Description of datasets and experimental setup}
\label{sec:er_setup}
~

The datasets of Table~\ref{tbl:6_datasets}, are used for the generation of the adversarial drifts. All data was normalized to the range of [0,1], and the data was converted to numeric/binary features type only. The synthetic datasets is a 10 dimensional dataset, with two classes. The \textit{Legitimate} class is normally distributed with a $\mu$-0.75 and $\sigma$=0.05, and the \textit{Malicious} class is centered at $\mu$-0.25 and $\sigma$=0.05, across all 10 dimensions. The CAPTCHA dataset is taken from \citep{d2014avatar}, and it represents an application concerned with the classification of mouse movement data for humans and bots, for the task of behavioral authentication. The phishing dataset is taken from \citep{Lichman:2013}, and it represents classification between malicious and benign web pages. The digits dataset \citep{Lichman:2013}, was taken to represent a standard classification task. Only classes 0 and 8 were considered, so as  to convert it to a binary classification task. In all datasets, the class 0 was considered to be the \textit{Legitimate} class. After the dataset is prepared, it is used in the \textit{AP-Stream} framework, to generate the adversarial stream. The data stream is considered balanced with a $\lambda_{Imbalance}$=0.5 and the blending ratio $\lambda_{Blend}$ is taken as 0.05, to avoid triggering drifts in the exploration phase. Detection uses a threshold of $\theta$=3 (based on analysis in \citep{sethi2017reliable}). The parameters of labeling and retraining $N_{train}$, and the profile of the stream are discussed at the beginning of each of the following subsections.

\begin{table}[t]
\centering
\caption{Description of datasets used for adversarial drift evaluation.}
\label{tbl:6_datasets}
\begin{tabular}{|c|c|c|}
\hline
Dataset & \#Instances & \#Attributes \\ \hline
Synthetic & 500 & 10 \\ \hline
CAPTCHA & 1886 & 26 \\ \hline
Phishing & 11055 & 46 \\ \hline
Digits08 & 1499 & 16 \\ \hline
\end{tabular}
\end{table}

The generation of adversarial drift is based on real world datasets, which are used in the \textit{AP-Stream} framework to simulate a streaming environment. The AP attacks are filtered as described in Figure~\ref{fig:er_filter}, to remove the low confidence samples. This filtering is done by using a random subspace model (50 Linear SVM, 50\% of features in each model), with a high regularization constant c=10, to ensure robustness against stray probes. Samples with confidence less than the confidence threshold of $\theta_{Adversary\_confidence}$=0.8 are eliminated, before exploitation starts. In all experiments in this section, averages are reported over 10 runs of the experiments. The experiments are performed using python and the scikit-learn machine learning library \citep{scikit-learn}.

\subsection{Analysis on data streams with a single simulated adversarial drift}
\label{sec:er_single}


\begin{figure*}[t]
\centering
\subfloat[synthetic]{\includegraphics[width=0.24\linewidth]{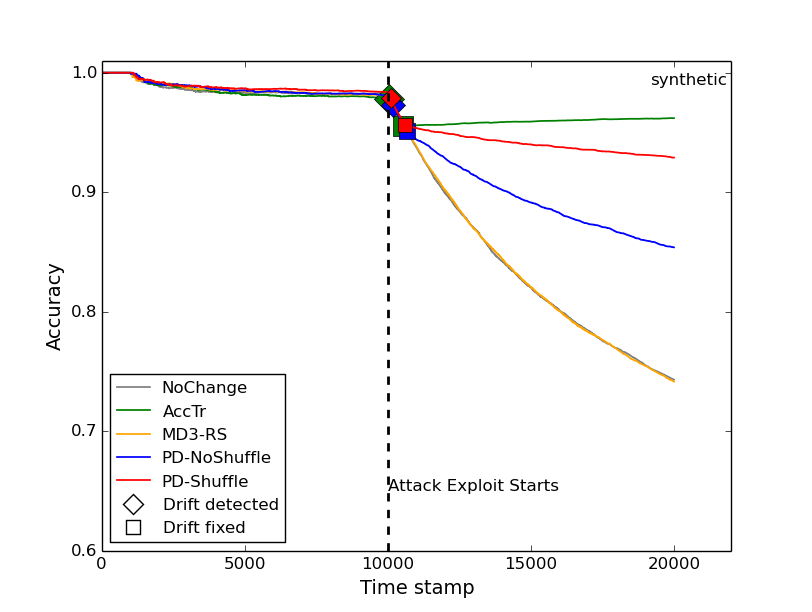}}
\subfloat[CAPTCHA]{\includegraphics[width=0.24\linewidth]{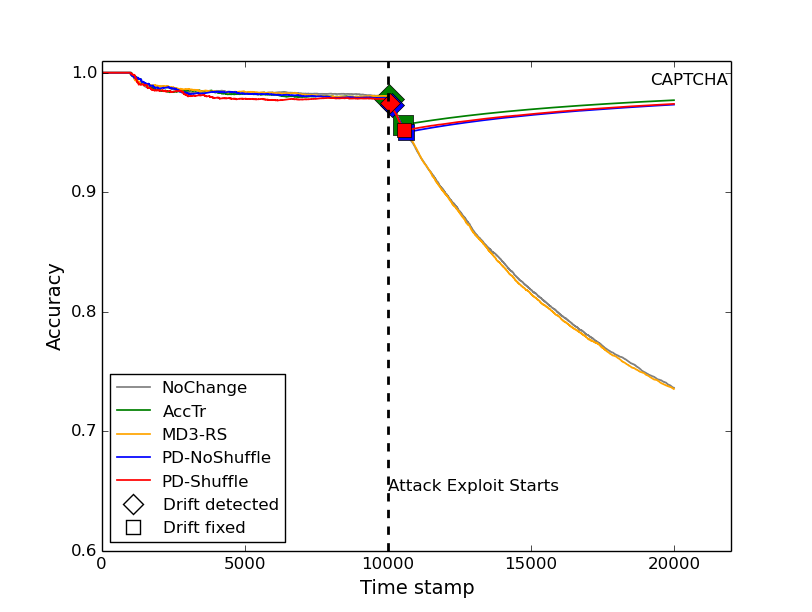}}
\subfloat[phishing]{\includegraphics[width=0.24\linewidth]{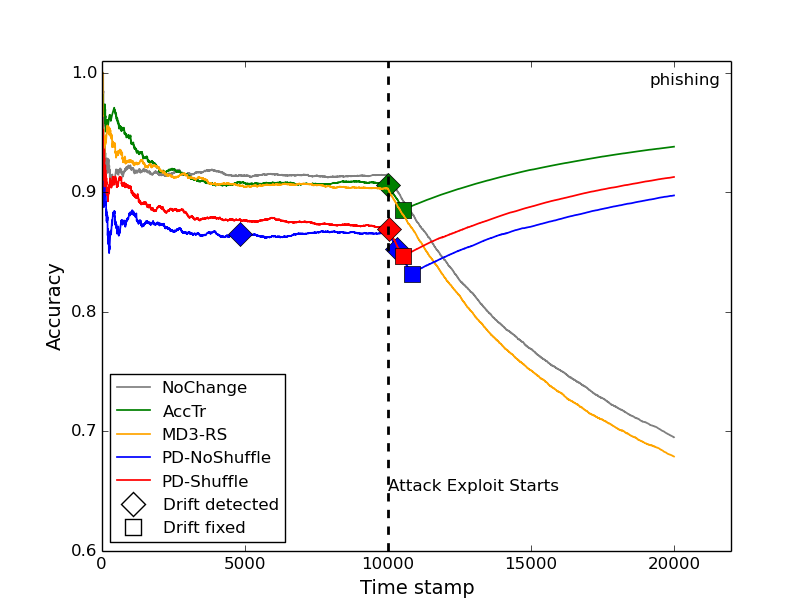}}
\subfloat[digits08]{\includegraphics[width=0.24\linewidth]{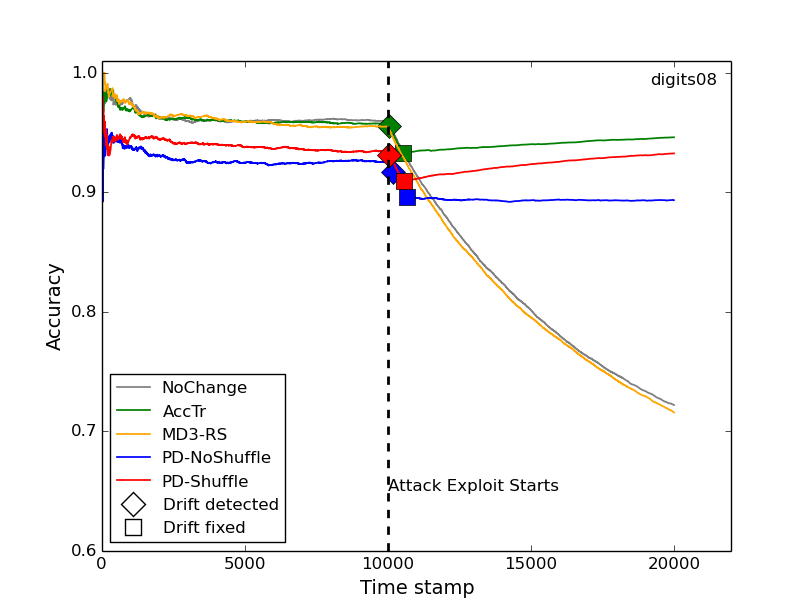}}
\caption{Accuracy over time for streams with single adversarial drift.}
\label{fig:er_single_acc}
\end{figure*}

\begin{figure*}[t]
\centering
\subfloat[synthetic]{\includegraphics[width=0.24\linewidth]{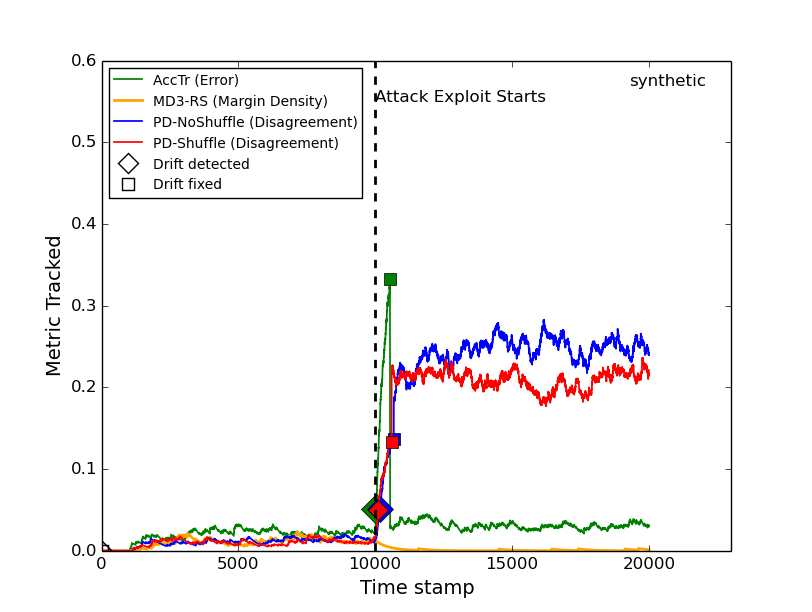}}
\subfloat[CAPTCHA]{\includegraphics[width=0.24\linewidth]{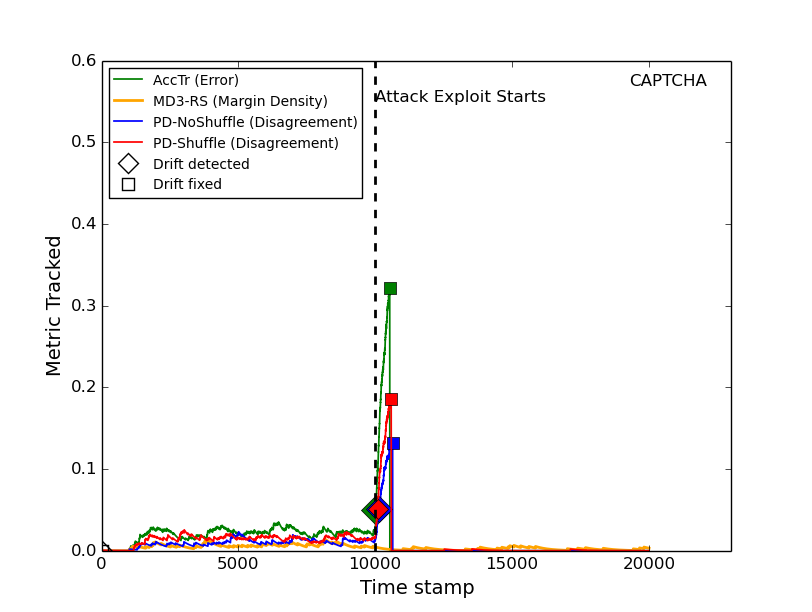}}
\subfloat[phishing]{\includegraphics[width=0.24\linewidth]{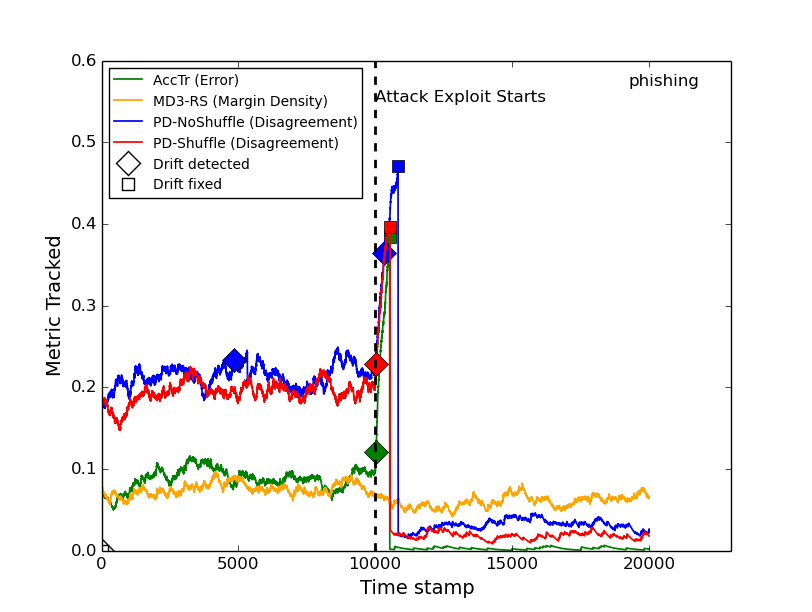}}
\subfloat[digits08]{\includegraphics[width=0.24\linewidth]{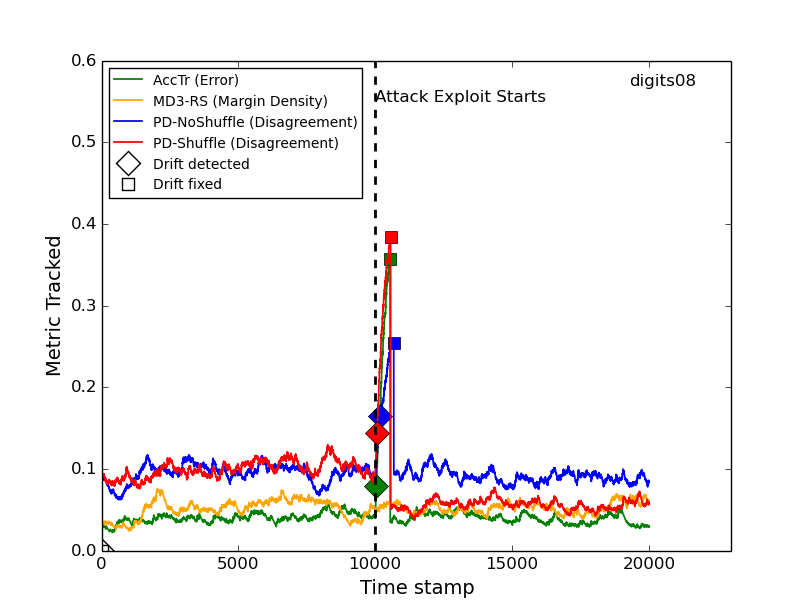}}
\caption{Metrics being tracked by the different drift handling techniques. AccTr tracks error rate based on fully labeled data. MD3 tracks margin density from unlabeled data. The PD-Shuffle and the PD-NoShuffle track the disagreement between the prediction model and the hidden detection model. }
\label{fig:er_single_metric}
\end{figure*}

In this section, we evaluate the performance of the different drift handling methodologies, on a data stream with a single adversarial drift. Drift is simulated using the \textit{AP-Stream} framework, with a stream length of 20,000 samples, a $t_{Explore}$=1000, and a $t_{Exploit}$=10000. The chunk size is taken to be $N$=500, and retraining/confirmation is performed by considering $N_{train}=N$ additional labeled samples, after drift indication.

The results of the Anchor Points (AP) attacks are shown in Figure~\ref{fig:er_single_acc}. It can be seen that the AP attacks, when used in the \textit{AP-Stream} simulator, leads to a concept drift which causes the performance to drop after $t_{Exploit}$=10,000. This is visible from the sudden drop in accuracy in the performance of the NoChange methodology, which assumes that the data will be static throughout. The drop in accuracy during the exploration phase (t=1,000-10,000) is minimal, in accordance with out simulation goals, as an adversary tries to hide its tracks while performing reconnaissance of the system. The need for drift detection and retraining is highlighted by the decreasing performance of the NoChange model, which rapidly becomes unusable after the drift. The fully labeled AccTr approach is able to effectively and quickly detect changes in the stream, allowing it to fix itself and maintain high accuracy, in all cases. This however comes at an unrealistic assumption of all labels being available immediately. Nevertheless, these two methodologies provide the upper and the lower baseline for our comparative analysis. 

\begin{table}[t]
\centering
\caption{Accuracy, Drift detected and Labeling\% of the NoChange, AccTr, MD3, PD-NoShuffle and PD-Shuffle methodologies, over streams with a single adversarial drift. }
\label{tbl:er_single}
\scalebox{0.65}{
\begin{tabular}{|c|l|c|c|c|}
\hline
Dataset & Methodology & Accuracy & Labeling\% & Drift Detected \\ \hline
\multirow{5}{*}{synthetic} & NoChange & 74.2 & 0 & No \\ \cline{2-5} 
 & AccTr & 95.7 & 100 & Yes \\ \cline{2-5} 
 & MD3 & 74.2 & 0 & No \\ \cline{2-5} 
 & PD-NoShuffle & 85.4 & 5 & Yes \\ \cline{2-5} 
 & PD-Shuffle & 93.7 & 2.5 & Yes \\ \hline
\multirow{5}{*}{CAPTCHA} & NoChange & 74.3 & 0 & No \\ \cline{2-5} 
 & AccTr & 97.7 & 100 & Yes \\ \cline{2-5} 
 & MD3 & 73.9 & 0 & No \\ \cline{2-5} 
 & PD-NoShuffle & 97.4 & 2.5 & Yes \\ \cline{2-5} 
 & PD-Shuffle & 96.5 & 2.5 & Yes \\ \hline
\multirow{5}{*}{phishing} & NoChange & 68.9 & 0 & No \\ \cline{2-5} 
 & AccTr & 93.8 & 100 & Yes \\ \cline{2-5} 
 & MD3 & 68.5 & 0 & No \\ \cline{2-5} 
 & PD-NoShuffle & 90.1 & 4.18 & Yes \\ \cline{2-5} 
 & PD-Shuffle & 91.3 & 2.5 & Yes \\ \hline
\multirow{5}{*}{digits08} & NoChange & 71.9 & 0 & No \\ \cline{2-5} 
 & AccTr & 95.6 & 100 & Yes \\ \cline{2-5} 
 & MD3 & 71.7 & 0 & No \\ \cline{2-5} 
 & PD-NoShuffle & 91.8 & 2.5 & Yes \\ \cline{2-5} 
 & PD-Shuffle & 92.9 & 2.5 & Yes \\ \hline
\end{tabular}}
\end{table}

The results of Figure~\ref{fig:er_single_acc} and Table~\ref{tbl:er_single}, demonstrate that the MD3 approach fails to detect adversarial drifts, as the accuracy of the MD3 approach is seen to be no better than the NoChange approach. Both approaches fail to detect drift. The filtered AP attack samples, evade several of the features simultaneously, causing the margin density based detection, to be circumvented. Margin density relies on having a few informative features drifting, while the other remain static \citep{sethi2017reliable}. This condition is intentionally violated by an adversary seeking to go undetected for a long time (Figure~\ref{fig:adverarial_drif_avoid}). As such, the generated drift samples fall outside the regions of uncertainty (margins), of the defender, leading to failed unsupervised attack detection. The PD approach was developed to address this issue, by using an adversarial aware design. It does so by hiding feature importance, thereby shielding them from probing based attacks. This ensures that the adversary will have a misguided notion of confidence, as it will not be able to obtain information about the exact influence of a subset of feature, no matter how high its exploration probing budget gets. By tracking the successful reverse engineering and evasion of a subset of exposed features, with an increased uncertainty over the other set of hidden features, the PD methodology aims to better detect adversarial activity. 

The \textit{Predict-Detect} framework is able to detect and recover from the adversarial drifts, shown by high performance after attacks, in Figure~\ref{fig:er_single_acc}. The drift detection is prompt and close to the AccTr approaches ($\Delta=3.3\%$, on average). The difference in performance, between the two PD approaches, is seen in case of the final accuracy, due to the different retraining approaches used. The PD-Shuffle outperforms PD-NoShuffle, due to the reassessment and re-splitting of feature importance, possible in the retraining phase. However, the PD-NoShuffle approach provides security against possible temporal attacks, in which an adversary might be able to collect information about different features over time, to cause a more potent data nullification attack. It does so with a compromise in the accuracy ($<$2.5\%, on average), compared to the PD-Shuffle methodology, which is a reasonable trade off. In either case, the PD methodologies are able to detect drifts with $<$5\% labeling, and are able to maintain classifier performance over time, as seen in Table~\ref{tbl:er_single}. The higher labeling budget for the PD-NoShuffle case is due to the increase in false alarms, which leads to requesting of additional labeled data that are eventually discarded as they indicate no significant drop in the accuracy. 

The progression of the metrics tracked by the different drift detectors is shown in Figure~\ref{fig:er_single_metric}. AccTr tracks the error rate (or accuracy), MD3 tracks the margin density, and the PD track the disagreement between the \textit{Prediction} and the \textit{Detection} models. It can be seen that the metrics of accuracy and the PD disagreement  depict a significant jump at the attack exploitation phase, and remain relatively stable before and after the drift. This indicates a high signal-to-noise ratio for these information metrics, and their effectiveness in detecting drifts, which are caused by adversarial attacks. It is also seen that the margin density metric is unable to detect any changes in the face of an attack, leading to its inefficacy. The adversarial aware PD approaches, are able to detect attacks similar to the AccTr approach. This demonstrates the importance of accounting for an adversarial aware design, in the training phase of a classifier, and the effectiveness of simple solutions implemented in the design of a classifier, leading to long term security benefits. The preemptive strategy of hiding feature importance, leads to benefits in terms of better attack detection, lower dependence on labeled data, and effective responsiveness to attacks, for higher availability and security, of the machine learning based system.

\subsubsection{Effects of varying the number of hidden features}
\label{sec:er_vary}

In the experiments so far, the available set of features are considered to be split equally split between the \textit{Prediction} and the \textit{Detection} model, based on the feature importance based splitting methodology of Section~\ref{sec:pm_framework_generate}. This was motivated by the intuition to consider prediction on incoming samples and detection of adversarial drifts, as equally important tasks, over the course of the stream. The main reason for the effectiveness of the \textit{Predict-Detect} design, is its ability to hide the feature importance of some of the features, from being probed and reverse engineered by an adversary. As such, it should be sufficient to hide fewer important features, in case they are sufficient to represent a high accuracy orthogonal representation of the training data. Generally, it is expected to have more features in the \textit{Prediction} model, as this will enable better predictive performance for normal functioning of the ML based service, and at the same time delay the onset of attacks. Here, we evaluate the effects of reducing the number of features included in the hidden detection classifier, and its impact on the detection and prediction capabilities of the framework.

\begin{table}[t]
\centering
\caption{Effects of varying the percentage of important hidden features, on the accuracy over the adversarial data stream.}
\label{tbl:varied_acc}
\scalebox{0.70}{

\begin{tabular}{|c|c|c|c|c|c|c|}
\hline
\multirow{2}{*}{\begin{tabular}[c]{@{}c@{}}Dataset /\\ \% of hidden features $\rightarrow$\end{tabular}} & \multicolumn{3}{c|}{PD-NoShuffle} & \multicolumn{3}{c|}{PD-Shuffle} \\ \cline{2-7} 
 & 10\% & 25\% & 50\% & 10\% & 25\% & 50\% \\ \hline
synthetic & 0.96 & 0.91 & 0.88 & 0.96 & 0.95 & 0.95 \\ \hline
CAPTCHA & 0.98 & 0.97 & 0.97 & 0.98 & 0.97 & 0.97 \\ \hline
phishing & 0.92 & 0.92 & 0.91 & 0.93 & 0.92 & 0.91 \\ \hline
digits08 & 0.96 & 0.95 & 0.94 & 0.96 & 0.95 & 0.94 \\ \hline
\end{tabular}}
\end{table}

\begin{table}[t]
\centering
\caption{Effects of varying the percentage of important hidden features, on the number of drifts signaled.}
\label{tbl:varied_drifts}
\scalebox{0.70}{
\begin{tabular}{|c|c|c|c|c|c|c|}
\hline
\multirow{2}{*}{\begin{tabular}[c]{@{}c@{}}Dataset /\\ \% of hidden features $\rightarrow$\end{tabular}} & \multicolumn{3}{c|}{PD-NoShuffle} & \multicolumn{3}{c|}{PD-Shuffle} \\ \cline{2-7} 
 & 10\% & 25\% & 50\% & 10\% & 25\% & 50\% \\ \hline
synthetic & 1 & 1 & 1 & 1 & 1 & 1 \\ \hline
CAPTCHA & 1 & 1 & 1 & 1 & 1 & 1 \\ \hline
phishing & 2 & 1 & 1 & 2 & 1 & 1 \\ \hline
digits08 & 3 & 1 & 1 & 3 & 1 & 1 \\ \hline
\end{tabular}}
\end{table}

Table~\ref{tbl:varied_acc} and Table~\ref{tbl:varied_drifts}, present the effects of reducing the number of hidden features from 50\% to 10\% and 25\%. Reducing features in the hidden model is done by blanking out additional features, in accordance to the methodology of Section~\ref{sec:pm_framework_generate}. The features are still distributed round robin, based on their informativeness to the classification task. From Table~\ref{tbl:varied_acc}, it can be seen that the accuracy of the stream is not significantly affected by reduction in the number of the hidden features, for both the PD-Shuffle and the PD-NoShuffle scenarios. The number of drifts detected are also seen to be similar, as seen in  Table~\ref{tbl:varied_drifts}. Additional false alarms are seen in case of 10\% hidden features. However, the effects of these false alarms are minimal, compared to the savings in the labeling obtained by using this unsupervised drift detector. As a guideline, 25\%-50\% is suggested for the number of important features in the hidden model. 

An important consideration for the applicability of the \textit{Predict-Detect} framework, is the presence of multiple orthogonal features in the training dataset, which can result in disjoint classifiers each with high accuracy of prediction. In such a scenario, the prediction and the detection models form a self monitoring scheme, for detecting adversarial activity. So long as the two models can be made disjoint (feature wise) and have high prediction performance; the number of hidden features, the type of the individual models, and the training mechanism used, does not have a significant impact on the effectiveness of the framework.

\subsection{Analysis on data streams with a multiple subsequent adversarial drifts}
\label{sec:er_multiple}

In this section, we evaluate the performance of the different drift handling techniques in a streaming environment with multiple subsequent adversarial drifts. This stream represents a scenario where the system gets attacked, the defender adapts, and then the adversary relaunches an attack on the newly deployed system. This represents a real world scenario, where protecting against the onset of attacks is not sufficient, and security is a cyclic process. Instead, the ability to detect attacks, fix them, and continue to provide services, is paramount. This emphasizes a reactive approach to security, with an understanding of the adversarial nature of the problem, leading to long term benefits. 

Multiple adversarial drifts are simulated using the \textit{AP-Stream} framework. We generate 5 subsequent adversarial drifts, to simulate 5 attack-defense cycles, each of 20,000 samples. Exploration occurs from $(1000+20000*i)$ to $(10000+20000*i)$ samples, where $i=0,1,…,4$ represents the cycle number. In each cycle, the adversary learns about the defender, using probes, and then launches a evasion attack on the defender's classifier. The defender is responsible for detecting the onset of the attacks, and retraining in the event of a confirmed attack. After 10,000 samples (of the previous attack exploitation step), the adversary is perceptive of the inefficacy of its attacks, due to the updated defender's model, and as such it relaunches its attack on the defender. This launch starts the new adversarial cycle, by probing the updated defender model, and launching evasion attacks against it. We do not emphasize details about how an adversary detects that the defender has retrained itself. Instead, it is assumed that an adversary gets perceptive of this change, as it receives \textit{Accept/Reject} feedback on its submitted probe samples. The focus of experimentation is on the defender's ability to detect and retrain in the face of multiple subsequent adversarial drifts.

\begin{figure*}[t]
\centering
\subfloat[synthetic]{\includegraphics[width=0.24\linewidth]{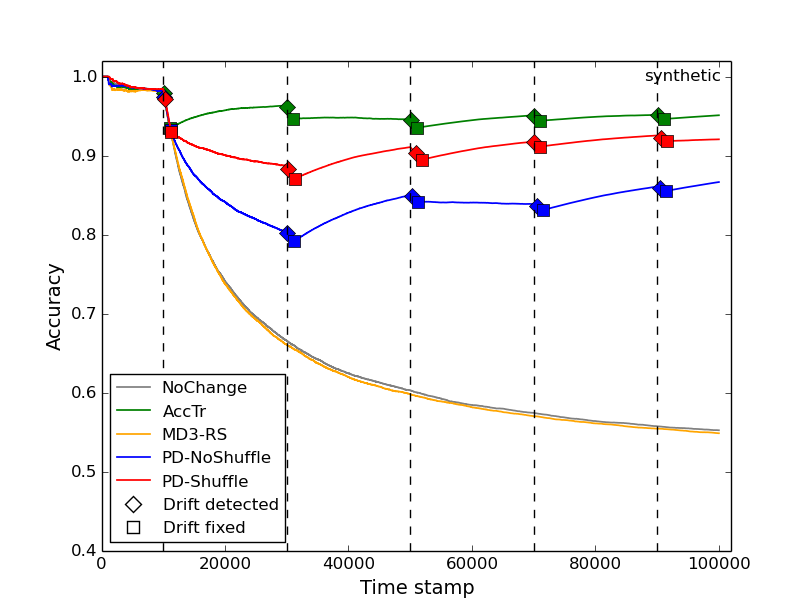}}
\subfloat[CAPTCHA]{\includegraphics[width=0.24\linewidth]{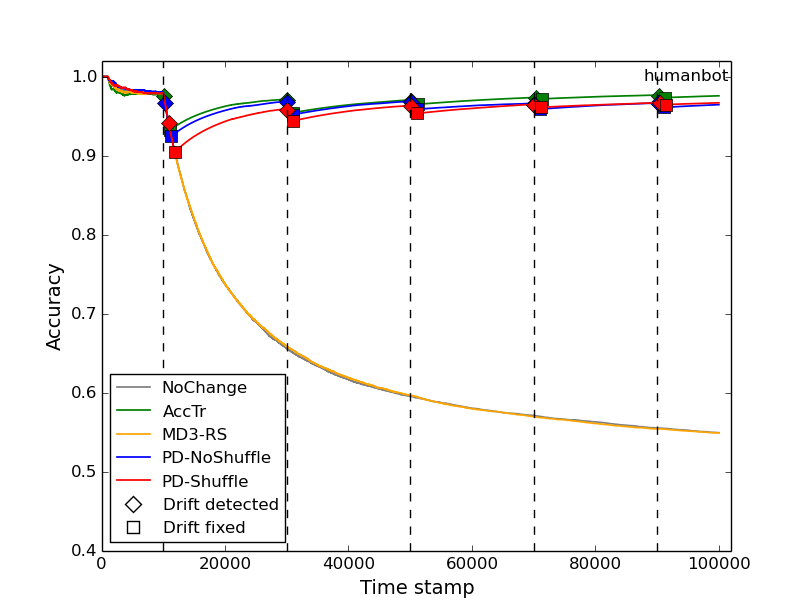}}
\subfloat[phishing]{\includegraphics[width=0.24\linewidth]{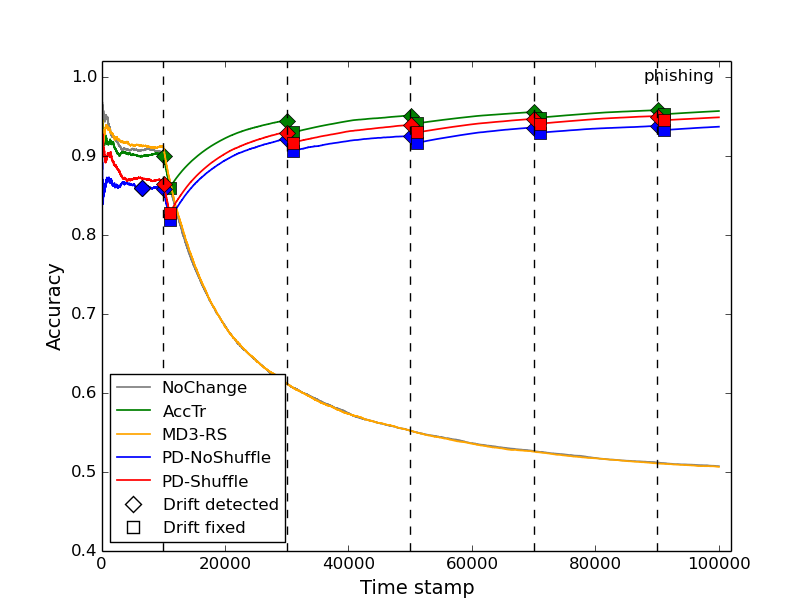}}
\subfloat[digits08]{\includegraphics[width=0.24\linewidth]{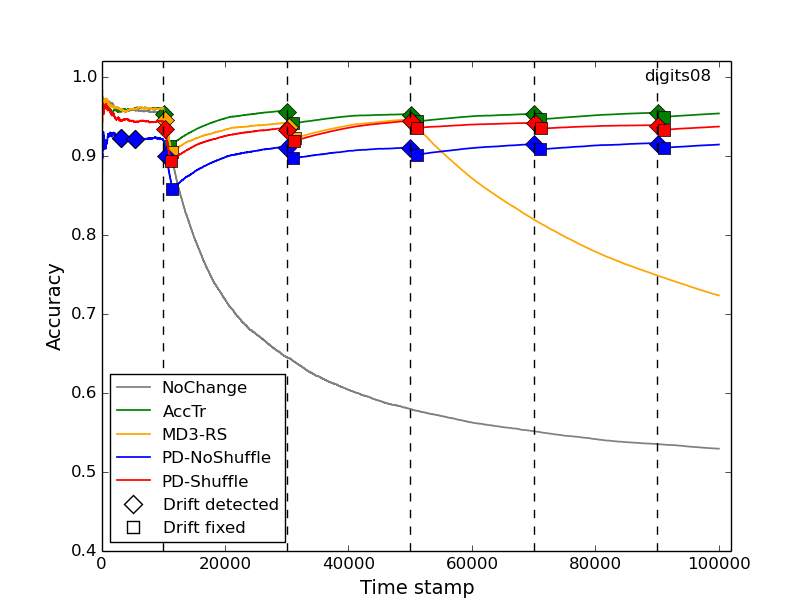}}
\caption{Accuracy over time for streams with multiple subsequent adversarial drifts.}
\label{fig:er_multiple_acc}
\end{figure*}

\begin{table}[t]
\centering
\caption{Results of drift handling, on data streams with multiple adversarial drifts.}
\label{tbl:multiple}
\scalebox{0.65}{

\begin{tabular}{|c|l|c|c|c|}
\hline
Dataset & Methodology & Accuracy & Labeling\% & Drifts Detected \\ \hline
\multirow{5}{*}{synthetic} & NoChange & 55.10 & 0 & 0 \\ \cline{2-5} 
 & AccTr & 94.80 & 100 & 5 \\ \cline{2-5} 
 & MD3 & 54.90 & 0 & 0 \\ \cline{2-5} 
 & PD-NoShuffle & 87.68 & 5 & 5 \\ \cline{2-5} 
 & PD-Shuffle & 90.54 & 5 & 5 \\ \hline
\multirow{5}{*}{CAPTCHA} & NoChange & 54.78 & 0 & 0 \\ \cline{2-5} 
 & AccTr & 97.57 & 100 & 4 \\ \cline{2-5} 
 & MD3 & 54.88 & 0 & 0 \\ \cline{2-5} 
 & PD-NoShuffle & 96.48 & 6 & 6 \\ \cline{2-5} 
 & PD-Shuffle & 96.77 & 5 & 5 \\ \hline
\multirow{5}{*}{phishing} & NoChange & 50.70 & 0 & 0 \\ \cline{2-5} 
 & AccTr & 95.66 & 100 & 5 \\ \cline{2-5} 
 & MD3 & 62.41 & 2 & 2 \\ \cline{2-5} 
 & PD-NoShuffle & 93.08 & 6 & 6 \\ \cline{2-5} 
 & PD-Shuffle & 94.65 & 5 & 5 \\ \hline
\multirow{5}{*}{digits08} & NoChange & 52.89 & 0 & 0 \\ \cline{2-5} 
 & AccTr & 95.03 & 100 & 5 \\ \cline{2-5} 
 & MD3 & 72.99 & 2 & 2 \\ \cline{2-5} 
 & PD-NoShuffle & 91.49 & 7 & 7 \\ \cline{2-5} 
 & PD-Shuffle & 93.53 & 5 & 5 \\ \hline
\end{tabular}}
\end{table}

The results over 5 cycles of the attack-defense cycle are presented in Table~\ref{tbl:multiple}, and the accuracy over the stream is depicted in Figure~\ref{fig:er_multiple_acc}. We consider a chunk size of $N$=1000, for all streams here, due to the increased size of the stream and the need for smoother analysis of the drift detection behavior. The progression of the streams in Figure~\ref{fig:er_multiple_acc}, indicates that the drift detection behavior of the methodologies is consistent with our observation for the single drift case. The MD3 approach fails to effectively detect and adapt to drifts, this is seen by the reduced accuracy of this approach when compared to the fully labeled drift detector AccTr ($34.5\%$  lower accuracy on average). This is a result of the inefficacy of the MD3 approach to track drifts caused by adversarial activity. This results in the MD3 approach being similar in performance to the NoChange methodology ($\Delta$=7.9\%). 

\begin{figure*}[t]
\centering
\subfloat[synthetic]{\includegraphics[width=0.24\linewidth]{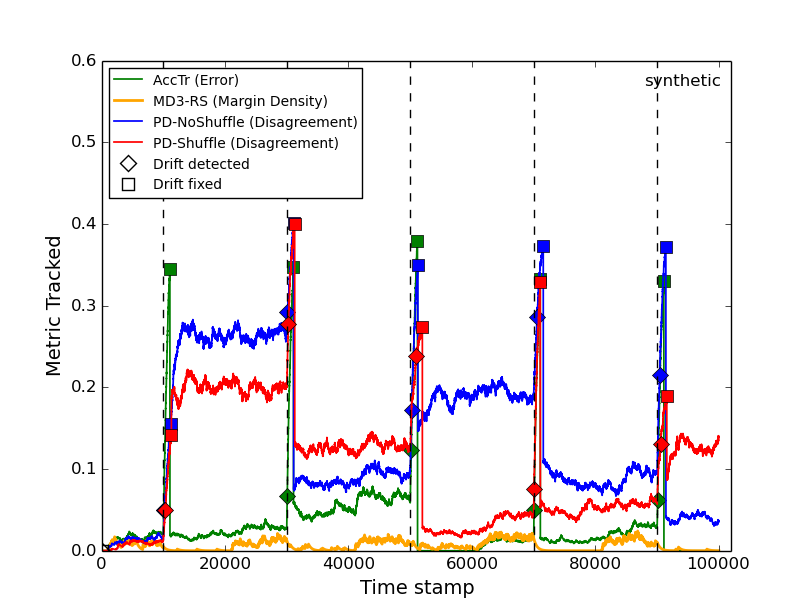}}
\subfloat[CAPTCHA]{\includegraphics[width=0.24\linewidth]{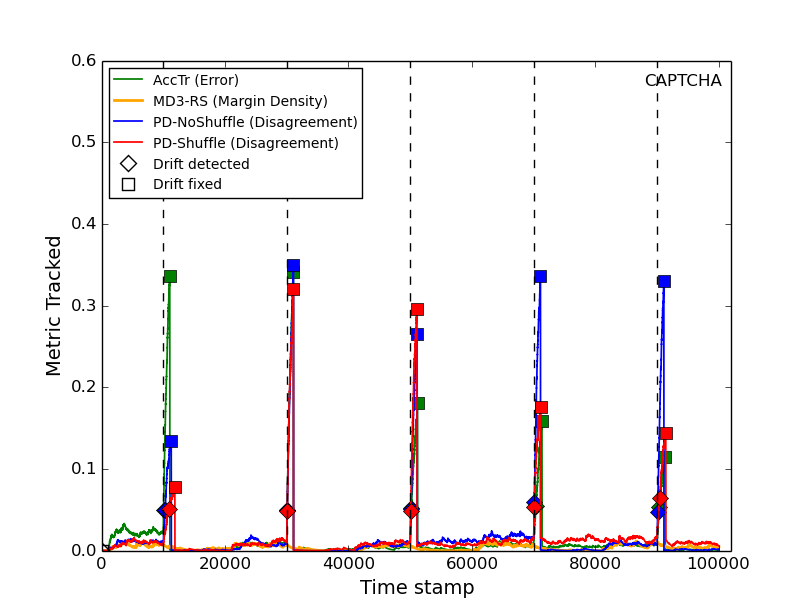}}
\subfloat[phishing]{\includegraphics[width=0.24\linewidth]{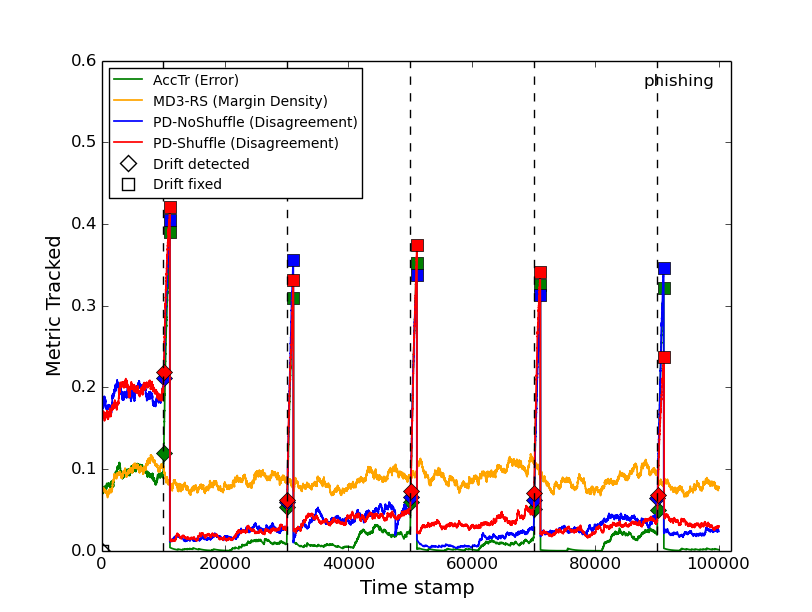}}
\subfloat[digits08]{\includegraphics[width=0.24\linewidth]{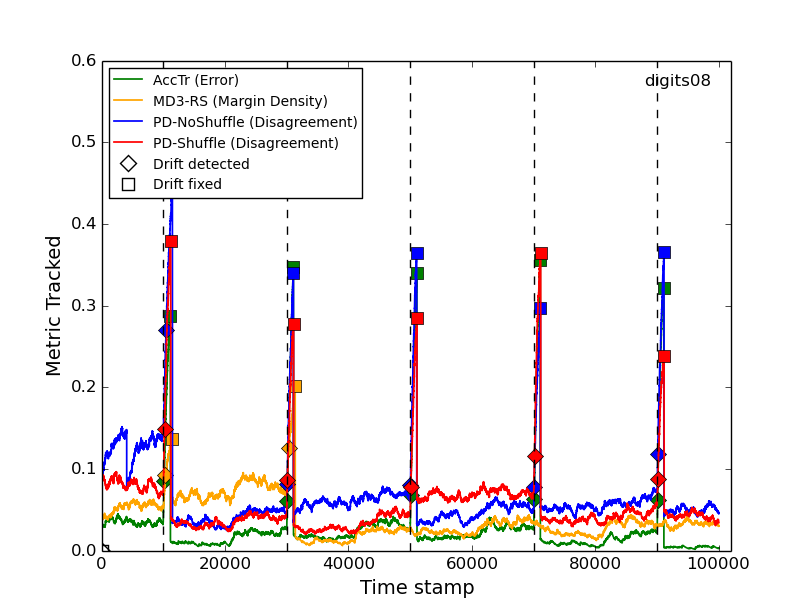}}
\caption{Metrics being tracked by the different drift handling techniques, for the data streams with multiple subsequent adversarial drifts. AccTr tracks error rate based on fully labeled data. MD3 tracks margin density from unlabeled data. The PD-Shuffle and the PD-NoShuffle track the disagreement between the prediction model and the hidden detection model.}
\label{fig:er_multiple_metric}
\end{figure*}

The PD methods provide performance similar to the fully labeled AccTr approach ($\Delta$=1.9 for PD-Shuffle and $\Delta$=3.6 for PD-NoShuffle), as it is able to effectively detect and recover from attacks in all cases. This is achieved at a low labeling rate of 6\% for the PD-NoShuffle and 5\% for the PD-Shuffle. This is because the \textit{Predict-Detect} framework provides a natural, adversarial-aware, unsupervised mechanism for detecting drifts. This minimum required labeling is needed, for drift confirmation and retraining. The drift detection metrics are depicted in Figure~\ref{fig:er_multiple_metric}. It is seen that the disagreement score for the PD serves as an effective surrogate to the fully labeled AccTr approach, as both the signals changes concurrently and with minimal lag, in the event of an attack. The high peak at the exploitation stage of attacks indicates the high signal-to-noise ratio of the disagreement tracking signal, and its effectiveness in detecting attacks. This makes the detectors resilient to stray changes, reserving labeling budget and model retraining only for cases where the adversarial attack can lead to serious degradation in the performance. The margin density signal is seen to be ineffective for detecting attacks, as it misses a majority of the drift scenarios in these experiments. This further highlights the need for an adversarial aware design when implementing streaming data algorithms for security applications. 

The PD-Shuffle performs marginally better than the PD-NoShuffle methodology, with an average improvement of 2\% in the accuracy of the stream. However, the PD-NoShuffle provides better protection by hiding features importance from an adversary, over time. The drift detection and retraining lag is similar for the PD and the AccTr cases, indicating the effectiveness of the detection capabilities of the proposed framework. This makes the \textit{Predict-Detect} framework effective for usage in an unsupervised, adversarial-aware, and streaming environment. The preemptive design of the framework makes it better prepared for dealing with adversarial activity, by enabling reliable unsupervised detection, and ensuring that recovering from drifts is a possibility.

\section{Disagreement based active learning on imbalanced adversarial data streams}
\label{sec:al}

The analysis thus far, has focused on the ability of the \textit{Predict-Detect} framework to detect drifts from unlabeled data, and subsequently recover from them. After a drift is indicated by the framework, labeling additional samples for drift confirmation and for retraining of models, is considered to follow naive strategies. More specifically, 100\% of samples from a chunk of $N_{Unlabeled}$ subsequent samples are requested to be labeled, and these samples are then used to confirm drifts and retrain the model. Although this approach is effective for a balanced stream, i.e., with near equal probability of occurrence of  the \textit{Legitimate} and the \textit{Malicious} class samples, it is not an efficient strategy for imbalanced data streams. This is especially pertinent in adversarial environments, as the adversarial attacks are expected to be a smaller minority class, in comparison to the majority of benign traffic entering the system. To improve the labeling process, we analyze the effect of selecting $N_{train}$ samples ($N_{train}<N_{Unlabeled}$) to label, for drift confirmation and retraining, after the framework indicates a drift from the unlabeled data. In our proposed active learning strategy, we use the motivation that samples falling in the disagreement region of the two classifiers have higher probability of belonging to the adversarial class. Section~\ref{sec:al_method} presents the proposed active learning methodology, based on the disagreement information between the \textit{Prediction} and the \textit{Detection} models. Experimental analysis is presented in Section~\ref{sec:al_er}.

\subsection{Active learning using disagreement information - \textit{Disagreement Sampling}}
\label{sec:al_method}

Adversarial drifts are detected by the \textit{Predict-Detect} framework, by tracking the disagreement between the two component models, over time. The efficacy of the detection mechanism relies on the inability of an adversary to successfully reverse engineer all of the feature information, due to the hidden inaccessible detection model. This causes adversarial samples to fall in the disagreement regions of the two classifier models. Here, the disagreement region serves as a honeypot, to capture adversarial samples. We extend this motivation to the domain of active learning, for selecting samples for labeling, after a drift is indicated. 

\begin{figure}[t]
  \centering
  \includegraphics[width=0.85\linewidth]{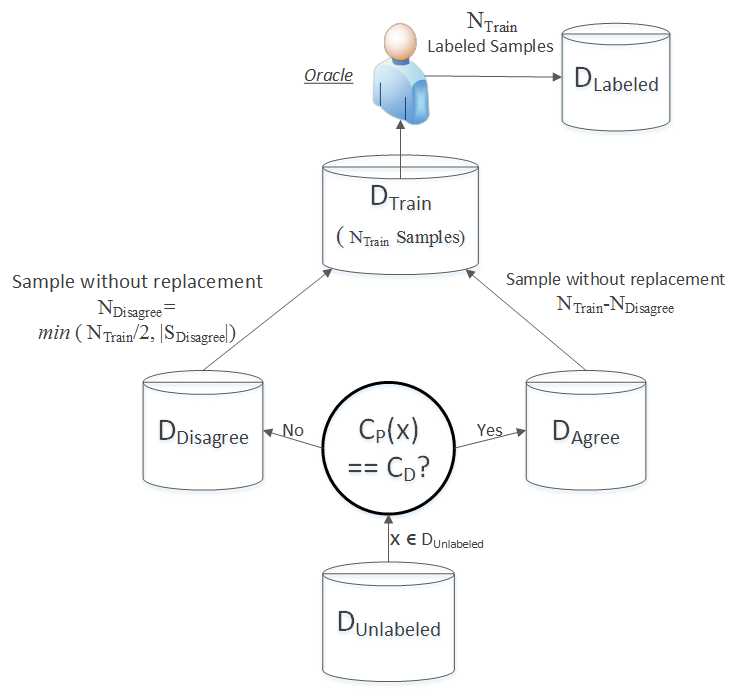}
   \caption{Disagreement based active learning methodology.}
  \label{fig:imbalance}
\end{figure}

The active learning based on disagreement is performed on the collected $N_{Unlabeled}$ samples, after the drift is indicated by the PD model. The proposed active learning process is shown in Figure~\ref{fig:imbalance}. The initial set of collected $N_{Unlabeled}$ unlabeled samples, after drift indication,  are processed, and $N_{train}$ samples are selected ($N_{train}<=N_{Unlabeled}$) to be labeled by the Oracle. The active learning methodology aims to embody maximum informativeness into the selected $N_{train}$ samples, about the newly drifted stream distribution. A large number of adversarial class samples are aimed to be selected, as this is the drifting class and is often the minority in most real world applications. A good number of adversarial class samples selected, will lead to more balanced datasets, for better drift confirmation and retraining of models.

The active learning methodology is presented in Figure~\ref{fig:imbalance}. The set of unlabeled data samples $N_{Unlabeled}$, collected after drift is indicated, is split into two pools: $D_{Agree}$ and $D_{Disagree}$, based on the predicted class labels of the \textit{Prediction} model  ($C_P$) and the \textit{Detection} model ($C_D$). Samples to be labeled are first obtained by random sampling on the pool of disagreeing samples $D_{Disagree}$, until the samples in this pool are exhausted, or we reach 50\% of our labeling budget ($N_{train}$). This ensures that we allocate more priority to the disagreement region, where we have a higher chance of obtaining the samples of the adversary class. The remaining samples are obtained by random selection on the samples which fall outside the disagreement region (i.e., by sampling from $D_{Agree}$). Since the data is predominantly assumed to be of the \textit{Legitimate} class, this sampling will allow us to receive balanced samples for better retraining on the attack samples, while providing secondary validation on the static nature of the legitimate class samples. The final set of $N_{train}$ samples is sent to the Oracle for labeling. By integrating the disagreement information into the labeling process, this methodology aims to achieve a more representative set of samples, for better detection and retraining over imbalanced adversarial streams. 

\subsection{Experimental analysis on imbalanced data streams with limited labeling}
\label{sec:al_er}

\begin{table*}[t]
\centering
\caption{Effects of \textit{Random Sampling}, on the classifier performance and the balance of the obtained re-training set, on the synthetic dataset. Italicized values indicate critical regions of evaluation – imbalanced stream with low labeling ratio. }
\label{tbl:rs_imbalance}
\scalebox{0.9}{
\begin{tabular}{|c|c|c|c|c|c|c|}
\hline
\multirow{2}{*}{\begin{tabular}[c]{@{}c@{}}Labeling$\rightarrow$ \\/ Imbalance$\downarrow$\end{tabular}} & \multicolumn{2}{c|}{100\%} & \multicolumn{2}{c|}{10\%} & \multicolumn{2}{c|}{5\%} \\ \cline{2-7} 
 & f-measure & Malicious\% & f-measure & Malicious\% & f-measure & Malicious\% \\ \hline
0.5 & 0.93 & 51.20\% & 0.92 & 52\% & 0.89 & 36\% \\ \hline
0.1 & 0.91 & 11.40\% & \textit{0.72} & \textit{10\%} & \textit{0.65} & \textit{8\%} \\ \hline
0.05 & 0.89 & 6.20\% & \textit{0.78} & \textit{4\%} & \textit{0.66} & \textit{4\%} \\ \hline
\end{tabular}}
\end{table*}

\begin{table*}[t]
\centering
\caption{Effects of \textit{Disagreement Sampling}, on the classifier performance and the balance of the obtained re-training set, on the synthetic dataset. Italicized values indicate critical regions of evaluation – imbalanced stream with low labeling ratio. }
\label{tbl:disagreement_imbalance}
\scalebox{0.9}{
\begin{tabular}{|c|c|c|c|c|c|c|}
\hline
\multirow{2}{*}{\begin{tabular}[c]{@{}c@{}}Labeling$\rightarrow$ \\/ Imbalance$\downarrow$\end{tabular}} & \multicolumn{2}{c|}{100\%} & \multicolumn{2}{c|}{10\%} & \multicolumn{2}{c|}{5\%} \\ \cline{2-7} 
 & f-measure & Malicious\% & f-measure & Malicious\% & f-measure & Malicious\% \\ \hline
0.5 & 0.92 & 51.60\% & 0.89 & 56\% & 0.91 & 61\% \\ \hline
0.1 & 0.90 & 9.00\% & \textit{0.86} & \textit{43\%} & \textit{0.86} & \textit{48\%} \\ \hline
0.05 & 0.94 & 5.20\% & \textit{0.88} & \textit{33\%} & \textit{0.84} & \textit{48\%} \\ \hline
\end{tabular}}
\end{table*}

Identifying and labeling samples of the adversarial class, in a streaming environment, is an important concern especially when the data stream is imbalanced. This is because, in a streaming data with drift in the minority class, the changes are harder to track and detect in the data space \citep{arabmakki2014rls}. In an adversarial environment, the attack class is often the minority class \citep{sculley2011detecting}, as opposed to the vast majority of legitimate samples being submitted to the system. As such, the adversarial class's impact could be shadowed by the legitimate class, causing the attacks to go unnoticed. We present experiments in this section, to provide an initial motivation into the usage of the disagreement samples, for better labeling and retraining of the models. 

\begin{figure}[t]
\centering
\subfloat[Effect of sampling, on f-measure of stream.]{\includegraphics[width=0.85\linewidth]{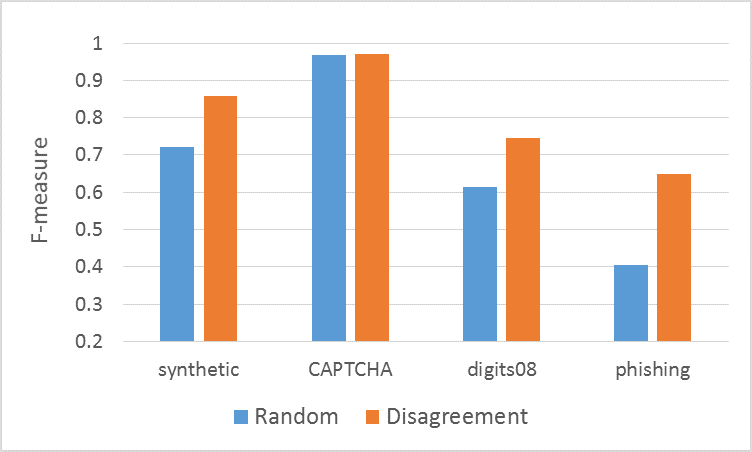}} \\
\subfloat[ Effect of sampling, on the percentage of malicious samples obtained for retraining. ]{\includegraphics[width=0.85\linewidth]{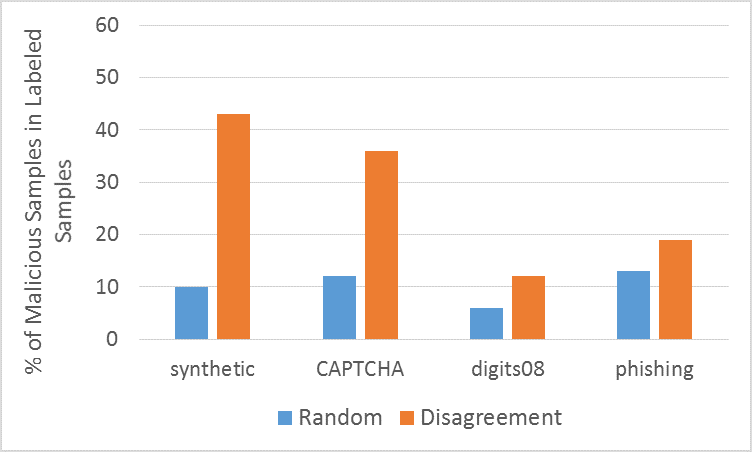}}  
\caption{: Effect of \textit{Random Sampling} and \textit{Disagreement Sampling}, at an imbalance rate of $\lambda_{Imbalance}$=0.1 and a labeling rate of 10\% (i.e., $N_{train}$=0.1 * $N_{Unlabeled}$).}
\label{fig:imbalance_compare}
\end{figure}

To demonstrate the effects of the proposed disagreement based active learning methodology, we first perform experiments on the synthetic dataset, using the same parameters and stream profile as in Section~\ref{sec:er_single} (20,000 samples, exploration starts at 1000 and exploitation starts at 10,000), and by using the \textit{AP-Stream} framework. The detection model is taken to be the PD model with shuffling of features for retraining (PD-Shuffle). The focus of this analysis is on the efficacy of the active learning approach based on the disagreement information, on the model retrain-ability, and the overall quality following the drift detection. Since the data stream is considered to be imbalanced, we update the Algorithm~\ref{algo:ad_detect}, to use f-measure for the confirmation of the drift,  instead of accuracy. The size of the $N_{Unlabeled}$ samples after drift detection is taken to be 500, and the detection threshold is kept at $\theta=$1.5, to account for imbalance in the data stream. The simulation of adversarial drifts in imbalanced streams is possible in the \textit{AP-Stream} framework, by usage of the $\lambda_{Imbalance}$ parameter.  The effect of varying the stream imbalance rate $\lambda_{Imbalance}$ and the labeling rate, is analyzed. Labeling rate is the fraction of samples in the set of $N_{Unlabeled}$ samples (buffered after drift indication), which will be labeled to form the set of $N_{train}$ samples, which are ultimately used for drift confirmation and retraining. 

The effect of varying the labeling rate and the imbalance in the data stream is shown in Table~\ref{tbl:disagreement_imbalance}. The f-measure of the final stream, and the percentage of malicious samples in the labeled $N_{train}$ samples, is shown in the table. High values on both metrics are desirable. As a baseline, the result of using random sampling (without replacement) is also presented in Table~\ref{tbl:rs_imbalance}. From Table~\ref{tbl:rs_imbalance} and Table~\ref{tbl:disagreement_imbalance}, we can observe that for imbalanced stream with limited labeling, the disagreement based active learning leads to a more balanced dataset for retraining. This results in a better f-measure for the stream, as more malicious samples are detected and retrained on. This is more pertinent in the case of reduced labeling (shown by emphasized values in both tables). Even for a 95\% imbalanced stream, and a budget of only 5\% of the $N_{Unlabaled}$ samples (=25 samples in this case), a f-measure of 0.84 is seen across the stream for the \textit{Disagreement Sampling}. This is beneficial, as the inability to detect the adversarial samples and train from them effectively, leads the random sampling approach to fall drastically in performance after attacks (with f-measure= 0.66).  The proposed PD model can detect drifts from imbalanced stream, and innately provides an active learning mechanism to further improve on labeling and retraining. This makes it attractive for usage in imbalanced streaming environments, where labeling is time consuming and expensive. 

The results of the proposed active learning methodology, on imbalanced streams with $\lambda_{Imbalance}$ of 0.1, and at a labeling budget of 10\% on the set of unlabeled training data, is shown in Figure~\ref{fig:imbalance_compare}. Across the 4 datasets, it can be seen that the active learning method outperforms the random labeling strategy by providing better f-measure and a more balanced labeled training dataset. While several active learning algorithms have been proposed in literature for streaming data \citep{zliobaite2014active}, the focus here is on the innate availability of information, due to the disagreement between the two model of the defender. This information is leveraged for drift detection from unlabeled data, and also for the task of active learning. The hidden classifier serves as a honeypot to capture adversarial samples, for our task of active learning and retraining. This design embodies the principle of dynamic adversarial learning, where forethought and adversarial awareness, leads to advantages in the detection and the subsequent retraining process.

\section{Discussion and extension of proposed framework}
\label{sec:discussion}

The developed \textit{Predict-Detect} framework, is used to demonstrate the benefits of feature importance hiding, in a streaming adversarial environment.  The information leverage on the defender's part, allows for advantages in reactive security, mainly for better attack detect-ability and recovery. In the design of the framework, it was assumed that the provided training dataset can be split into two orthogonal subsets, with each set providing sufficient predictive capabilities. In this section, we empirically evaluate this assumption on popular cybersecurity datasets, and also provide initial ideas for extensibility and customize-ability of the framework.

\begin{table*}[t]
\centering
\caption{Results of splitting datasets, vertically, into orthogonal feature subsets. Number of splits is given as $K$. Splits possible within 5\% and 10\% accuracy loss, are italicized.}
\label{tbl:extension}
\scalebox{0.9}{
\begin{tabular}{|l|c|c|c|c|c|}
\hline
\multicolumn{1}{|c|}{Dataset} & CAPTCHA & Phishing & Spam & Spamassassin & Nsl-Kdd \\ \hline
$\#$Instances & 1885 & 11055 & 6213 & 9324 & 37041 \\ \hline
$\#$Attributes & 26 & 46 & 499 & 499 & 122 \\ \hline
\begin{tabular}[c]{@{}l@{}}Monolithic Model \\ Accuracy\% (K=1)\end{tabular} & 99.9 & 96.3 & 96.2 & 97.5 & 98.1 \\ \hline
Min Accuracy\% at K = 2 & 99.8 & 92.1 & 95.9 & 96.7 & 97.1 \\ \hline
Min Accuracy\% at K = 3 & 99.6 & 88.2 & 95.4 & 96.2 & 92.1 \\ \hline
\begin{tabular}[c]{@{}l@{}}Max Partitions within \\ 5\% accuracy drop (K@5\%)\end{tabular} & \textit{5} & \textit{2} & \textit{6} & \textit{7} & \textit{2} \\ \hline
\begin{tabular}[c]{@{}l@{}}Model performance at  K@5\% \\ (Mean Accuracy, Min  Accuracy)\end{tabular} & \begin{tabular}[c]{@{}c@{}}(0.99, \\ 0.99)\end{tabular} & \begin{tabular}[c]{@{}c@{}}(0.92, \\ 0.92)\end{tabular} & \begin{tabular}[c]{@{}c@{}}(0.94, \\ 0.92)\end{tabular} & \begin{tabular}[c]{@{}c@{}}(0.94, \\ 0.93)\end{tabular} & \begin{tabular}[c]{@{}c@{}}(0.97, \\ 0.97)\end{tabular} \\ \hline
\begin{tabular}[c]{@{}l@{}}Max Partitions within \\ 10\% accuracy drop (K@10\%)\end{tabular} & \textit{9} & \textit{3} & \textit{10} & \textit{10} & \textit{4} \\ \hline
\begin{tabular}[c]{@{}l@{}}Model performance at,K@10\%\\  (Mean Accuracy, Min Accuracy)\end{tabular} & \begin{tabular}[c]{@{}c@{}}(0.99, \\ 0.95)\end{tabular} & \begin{tabular}[c]{@{}c@{}}(0.9, \\ 0.88)\end{tabular} & \begin{tabular}[c]{@{}c@{}}(0.90, \\ 0.89)\end{tabular} & \begin{tabular}[c]{@{}c@{}}(0.92, \\ 0.91)\end{tabular} & \begin{tabular}[c]{@{}c@{}}(0.94, \\ 0.93)\end{tabular} \\ \hline
\end{tabular}}
\end{table*}

Effects of splitting popular cybersecurity datasets is shown in Table~\ref{tbl:extension}. Experiments are performed on 5 datasets from the cybersecurity domain, namely: CAPTCHA \citep{d2014avatar}, phishing \citep{Lichman:2013}, spam, spamassassin \citep{katakis2010tracking,katakis2009adaptive} and nsl-kdd\citep{tavallaee2009detailed} dataset. The datasets were split vertically (i.e., based on features, with each feature containing all instances of the dataset), based on the feature ranking criterion of Algorithm~\ref{algo:generate_pd}. Each model in the table is a Linear SVM. Accuracy was computed using 10-fold cross validation over the entire dataset.

In Table~\ref{tbl:extension}, effects of splitting a dataset, based on features, into $K$ orthogonal subsets, is demonstrated.  For $K$=1, there is no split in the feature space and the learned model is trained on the entire original dataset. For $K>1$, the minimum accuracy of the $K$ trained models is reported. Additionally, the number of splits- $K$ possible at a 5\% and 10\% loss in the minimum accuracy, is also reported. From the results, it can be seen that it is reasonable to divide the dataset's features into $K>3$ sets of independent disjoint subsets, such that the minimum accuracy over any feature subset is within an acceptable range ($<$10\%), when compared to a monolithic classifier trained on the entire set of features. For the high dimensional datasets of spam and spamassassin, it is seen that 10 partitions are possible within 10\% accuracy of the monolithic classifier. We use minimum accuracy of the models for evaluation, as we are interested in feature set partitions, such that all splits have high information content. The results of Table~\ref{tbl:extension}, highlight the ability to split the high dimensional cybersecurity datasets into multiple orthogonal subsets of informative features. This evaluation also demonstrates that the split of the datasets in the \textit{Predict-Detect} framework, into only two subsets, is conservative, as the datasets can be effectively split into a larger number of subsets. 
 
The presence of orthogonal information, and the ability to train models on disjoint feature subsets, allows for extensibility of the \textit{Predict-Detect} framework. The framework can be extended to be a modular multiple classifier system (MCS) \citep{wozniak2014survey,farid2013adaptive,gomes2017survey}, which effectively leverages the orthogonal trained models.  Instead of using two classifiers - one for the \textit{Prediction} model and the other for the \textit{Detection} model, the framework can use two sets of ensembles - the \textit{Prediction Ensemble} and the \textit{Detection Ensemble}, each comprised of individual orthogonal models (Table~\ref{tbl:extension}). This allows system designers to leverage the benefits of feature importance hiding, while still providing flexibility of using their existing model designs and system architectures. The \textit{Prediction Ensemble} provides for the ability to apply existing robust learning methodologies of \citep{wang2015robust,biggio2010multiple,colbaugh2012predictability,alabdulmohsin2014adding}. While the \textit{Detection Ensemble} provides for the ability to maintain strategic advantages in a reactive environment. Presence of multiple models can provide better retraining and replacement policies, which are well designed for ensemble systems \citep{gomes2017survey}. 

The proposed framework can be made automated, where the feature splitting, retraining, and replacing models, can be done dynamically by the system, and the system is capable of requesting human intervention when it deems necessary. This would make it an intelligent dynamic system, which is capable of reacting to adversarial activity.  The domain agnostic design of the framework, assumes a black box view of the defender, further allowing for application specific customization, by system designers.

\section{Conclusion and future work}
\label{sec:conclusion}

In this paper, the characteristic of adversarial drift, as a special category of concept drift, is presented and analyzed. Adversarial concept drift leads to non-stationarity in the data mining process, and causes deployed machine learning models to degrade over time. It is necessary to deal with these concept drift, in the presence of limited labeling, for practical and continued usage of machine learning, in adversarial domains. As such, we introduce the \textit{Predict-Detect} streaming classification framework, which is capable of preemptively accounting for adversarial activity, thereby providing benefits at test time. By including considerations for unsupervised drift indication, selective labeling of samples, and misleading of attackers; the proposed framework provides for a generalized way for developers and data scientist, to design for long term security, in adversarial environments. 

The developed \textit{Predict-Detect} framework, was shown to outperform traditional unsupervised drift detection techniques (particularly MD3); in terms of drift detection, and recover-ability from the effects of an adversarial drift. Experimental evaluation was performed on data streams exhibiting adversarial concept drift, which were generated using a novel proposed simulation framework for adversarial drift. The adversarial agnostic design of MD3, was found to be vulnerable to evasion at test time, leading to its inability to detect and recover form drifts. On the other hand, the proposed \textit{Predict-Detect} framework was seen to be able to perform well in such environments, by its ability to account for adversarial activity in the drift detection process. The proposed framework was also seen to provide a savings of $>$94\% in the labeling budget, when compared to a fully labeled accuracy tracking concept drift handling approach, while maintaining similar accuracy over the stream ($\Delta <$4\%). In addition, the \textit{Predict-Detect} design of classifiers, provides for an innate ability to drive active learning on imbalanced data streams, as it serves as a honeypot for capturing minority class adversarial samples. The addition of the adversarial context to the design of the classifiers, at training time,  was seen to benefit the dynamic aspects of the system, a test time. This serves to motivate future work in the domain of adversarial drift handling, where forethought and appreciation of domain specific limitations, can provide for better designed systems. 

This work primarily focused on the benefits of adding adversarial awareness to the drift detection and retraining component or stream classification systems. To account for a holistic view of the adversarial landscape in which machine learning models operate, it is necessary to integrate such awareness into various aspects of the dynamic system design. Effects of causative attacks \citep{barreno2006can}, to mislead the retraining of the streaming data models, warrants further research. Also, the ability of an adversary to influence the labeling component of the framework, needs analysis. Ideas along the work of adversarial drift \citep{kantchelian2013approaches}, and adversarial active learning \citep{miller2014adversarial}, are initial steps towards a comprehensive solution.





\section*{References}

\bibliographystyle{elsarticle-harv} 
\bibliography{references}




\end{document}